\pdfoutput=1

\documentclass[runningheads]{llncs}

\usepackage{eccv}

\usepackage{eccvabbrv}

\usepackage{graphicx}
\usepackage{amsmath}
\usepackage{amssymb}
\usepackage{booktabs}
\usepackage[symbol]{footmisc}

\usepackage[accsupp]{axessibility}  %

\usepackage[pagebackref,breaklinks,colorlinks,citecolor=eccvblue]{hyperref}

\usepackage{orcidlink}
\usepackage{wrapfig}
\usepackage{graphicx}
\usepackage{amssymb}
\usepackage{amsmath}
\usepackage{booktabs}
\usepackage{bbm}
\usepackage{times}
\usepackage{epsfig}
\usepackage{tabu}
\usepackage{multirow}
\usepackage{amsfonts}       %
\usepackage{xpunctuate}
\usepackage{pifont}%
\usepackage{color,colortbl}
\usepackage{tabulary,overpic}
\usepackage{booktabs}       %
\usepackage[normalem]{ulem}

\usepackage{amsmath}
\usepackage{cite}

\usepackage{scalerel}
\usepackage[accsupp]{axessibility}

\usepackage{xcolor}
\usepackage{enumitem}
\usepackage{arydshln} %
\usepackage{xspace}
\usepackage{booktabs}
\usepackage{colortbl}
\usepackage{multirow}
\usepackage{makecell}
\usepackage{lipsum} %
\usepackage{gensymb} %

\usepackage{tabularx}

\definecolor{fyxcolor}{RGB}{0,128,255}
\definecolor{demphcolor}{RGB}{100,100,100}
\definecolor{citecolor}{RGB}{0,0,192} %
\definecolor{GrayBG}{gray}{0.95}

\newcommand{\app}{\raise.17ex\hbox{$\scriptstyle\sim$}}
\newlength\savewidth

\newrobustcmd{\B}{\bfseries}

\usepackage[labelsep=period]{caption}
\renewcommand{\paragraph}[1]{\vspace{0.2em}\noindent \textbf{#1 \hspace{0.2em}}}
\definecolor{lightcyan}{rgb}{0.88, 1.0, 1.0}

\definecolor{MyDarkRed}{rgb}{0.66, 0.16, 0.16}
\definecolor{MyDarkBlue}{rgb}{0.16, 0.16, 0.66}

\newcommand{\ours}{\textsc{sigma} }

\begin{document}

\title{\ours\!\!: Sinkhorn-Guided Masked Video Modeling
} 

\titlerunning{\ours\!\!: Sinkhorn-Guided Masked Video Modeling}

\author{Mohammadreza Salehi\textsuperscript{*}, Michael Dorkenwald\textsuperscript{*}, Fida Mohammad Thoker\textsuperscript{* \dag}, \\
Efstratios Gavves, Cees G. M. Snoek, Yuki M. Asano
}
\authorrunning{Mohammadreza Salehi et al.}

\institute{University of Amsterdam}

\maketitle

\footnotetext{\textsuperscript{*} denotes equal contribution. \textsuperscript{\dag} now at KAUST.} 

\begin{abstract}
Video-based pretraining offers immense potential for learning strong visual representations on an unprecedented scale. 
Recently, masked video modeling methods have shown promising scalability, yet fall short in capturing higher-level semantics due to reconstructing predefined low-level targets such as pixels. 
To tackle this, we present Sinkhorn-guided Masked Video Modelling \mbox{(\ours\!\!)}, a novel video pretraining method that jointly learns the video model in addition to a target feature space using a projection network. However, this simple modification means that the regular L2 reconstruction loss will lead to trivial solutions as both networks are jointly optimized. As a solution, we distribute features of space-time tubes evenly across a limited number of learnable clusters. By posing this as an optimal transport problem, we enforce high entropy in the generated features across the batch, infusing semantic and temporal meaning into the feature space. The resulting cluster assignments are used as targets for a symmetric prediction task where the video model predicts cluster assignment of the projection network and vice versa. 
Experimental results on ten datasets across three benchmarks validate the effectiveness of \ours in learning more performant, temporally-aware, and robust video representations improving upon state-of-the-art methods. Our project website with code is available at: \url{https://quva-lab.github.io/SIGMA}.
\end{abstract}

\section{Introduction}
Video-based pretraining offers an unprecedented scale of training data. With a simple estimate of 24 frames per second, 196 tokens per frame, and around 90K years of videos~\cite{youtube}, YouTube alone would contain around 10,000 trillion tokens. 
Compare this to the language domain where, for example, large-scale pretraining such as Llama-2~\cite{llama2} utilized ``only'' 2 trillion tokens. 
Videos offer not only a vast scale for pretraining, but also allow for understanding how objects interact with one another and transform across time. Such semantic temporal and spatial understanding is valuable for many applications ranging from autonomous driving~\cite{kuutti2020survey,chen2023end, janai2020computer} to robotic planning~\cite{finn2017deep, wellhausen2019should}, embodied AI~\cite{duan2022survey}, and learning world models~\cite{ha2018world, matsuo2022deep}.

As in the language domain, the key to achieving scalability lies in utilizing self-supervised learning signals. 
Initial works on self-supervised video learning focused on solving pretext tasks such as temporal ordering \cite{fernando2017self} or playback speed prediction \cite{yao2020video}. Those methods achieved relatively good performance on small-scale datasets like UCF-101, yet, these methods do not scale well with more pretraining data. More recently, the Vision Transformer (ViT)~\cite{dosovitskiy2020image} has shown to scale well with data and parameter size \cite{dehghani2023scaling, singh2023effectiveness} and has been successfully adapted to videos \cite{mvit_v1_fan, arnab2021vivit}, albeit, initially requiring large-scale labeled data. 

At present, self-supervised works utilizing a masked video modeling framework for vision transformers -- operating similarly to the `filling-in-the-blank' objective utilized in large language models -- now dominate the field and obtain state-of-the-art performances~\cite{videomae, guided_masking_fan, feichtenhofer2022masked}.
In this framework, part of the input data is masked and the network is trained to reconstruct predefined targets such as pixel values \cite{videomae, guided_masking_fan, wang2023videomaev2}, frame differences \cite{cmae_v_lu} or a Histograms of Oriented Gradients \cite{feichtenhofer2022masked}. 
While scalable and simple, these current methods do not fully capture the higher-level temporal and spatial semantics, as we also show empirically in this paper. 
This is because individual patches or space-time tubes do not represent individual semantic units, while in the language domain, words or subwords do. Therefore, tasking the model to reconstruct them pushes it towards learning low-level features.
Based on this, we propose a novel approach for masked video modeling that focuses on reconstructing more abstract temporal and spatial video representations.

\begin{wrapfigure}{r}{0.4\textwidth} 
\vspace{-2em}
    \includegraphics[width=0.4\textwidth]{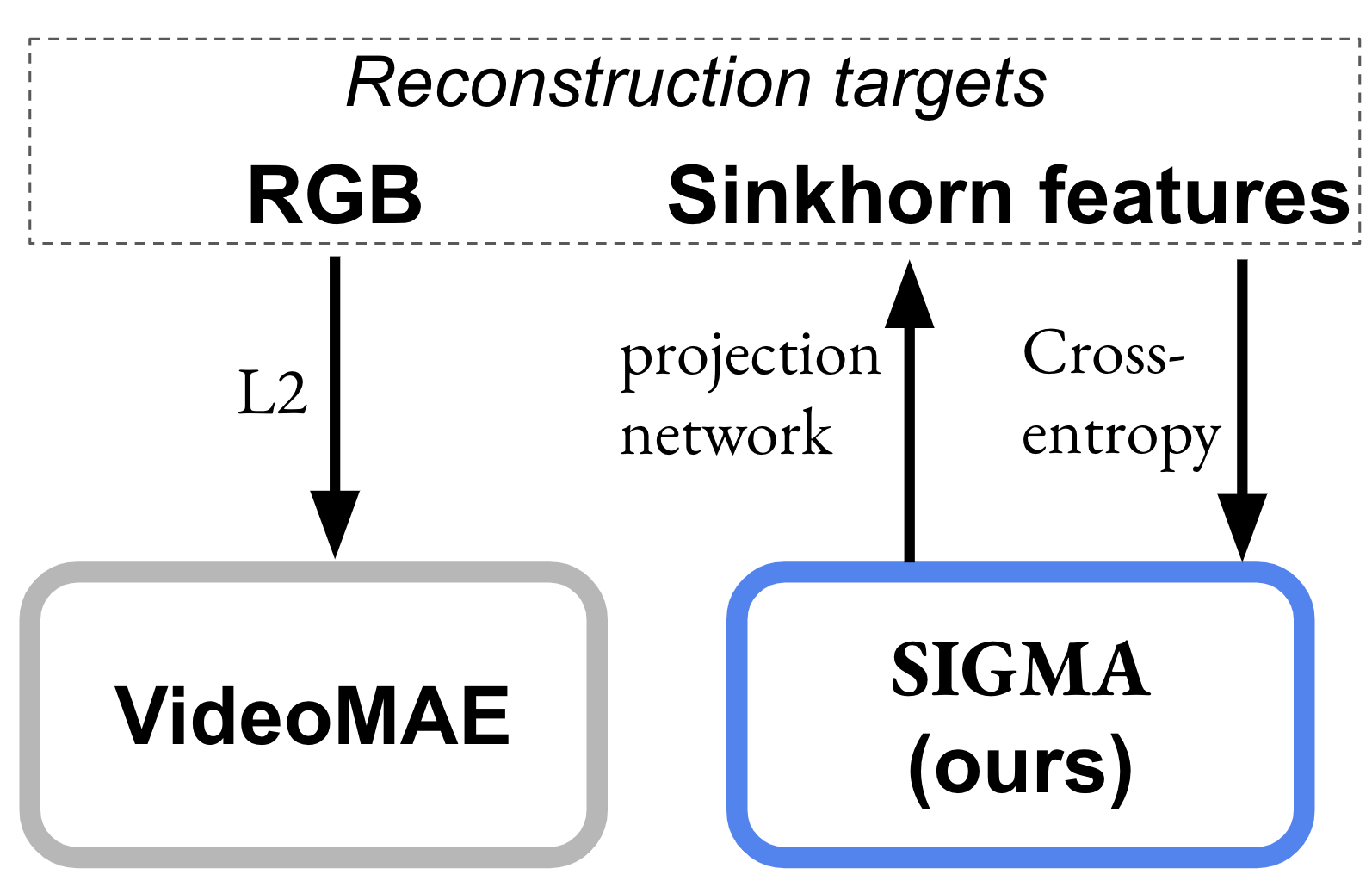}
    \caption{\textbf{Overview of our idea.} Compared to VideoMAE, which uses RGB pixels as targets, we generate Sinkhorn-regularised features as reconstruction targets. This obtains more semantic features and yields better pretraining performance.}
    \label{fig:fig1}
    \vspace{-2em}
\end{wrapfigure}

In this work, we propose a new framework wherein the typically predefined reconstruction target space can be simultaneously learned alongside the video model. 
For this, a projection network is introduced which embeds both the visible and masked portions of the video, yielding deep feature reconstruction targets. However, employing a commonly used L2 reconstruction loss naïvely is ineffective due to the joint optimization of both networks leading to a trivial solution as both networks collapse to the same output irrespective of the input.
To solve this, we introduce \textsc{sigma}: Sinkhorn-guided masked video modeling, where deep features of space-time tubes are regularised by optimal transport uniformly across clusters.
This effectively acts as a high-entropy regularization constraint and enforces similar space-time tube features to be assigned to the same centroid, infusing semantic meaning into the feature space.
These cluster assignments and centroids are learned in an online manner using the fast Sinhkhorn-Knopp algorithm, yielding feature pseudo-labels as targets. 
With these targets, we formulate our loss objective as a symmetric prediction task, where the features from each branch -- the video model and the projection network -- cross-predict the cluster assignment of the other. By doing so, we force the features of space-time tubes to be expressed by a limited number of clusters, enforcing semantic-rich concepts, while eliminating the dependency on predefined targets such as the masked pixel values, commonly used in prior works. Moreover, despite our cross-prediction task, we do not rely on any augmentations~\cite{qian2021spatiotemporal, patrick2020multi, dorkenwald2022scvrl} or crops~\cite{bardes2023mcjepa, mme_sun}, making our model stable and easy to train.

We conduct experiments on ten datasets across three benchmarks and demonstrate the superiority of our approach compared to the baseline VideoMAE~\cite{videomae}, as well as current state-of-the-art methods. 
We show that in a frozen evaluation setting our method outperforms previous works considerably on various video and even image datasets showcasing better transferability. Moreover, we evaluate the object-level understanding of our method, by evaluating its performance on an unsupervised video object segmentation benchmark~\cite{salehi2023time}, demonstrating better spatial and temporal semantics.
Finally, we evaluate the generalization abilities on the SEVERE benchmark~\cite{thoker2022severe} and improve upon existing works.
These findings confirm that \textsc{sigma} learns more performant, temporally-aware and robust representations.
We will make the code and models available.

Our contributions can be summarized as follows:
\begin{itemize}
    \item We propose \textsc{sigma}: a masked video modeling framework in which the reconstruction target space is jointly learned along with the video model.
    \item We extensively evaluate our pretraining method across standardized sets of datasets and benchmarks and obtain state-of-the-art results. 
    \item We highlight the flexibility of our approach by including DINO-pretrained features as reconstruction targets, which obtains further gains. 
\end{itemize}

\section{Related Works}

\noindent\textbf{Self-supervised Video Representation Learning.}
Self-supervised learning aims to learn features that can discriminate within an input distribution without the need for supervision. This is often achieved in images, by training models to accomplish pretext tasks, as outlined in works like \cite{noroozi2016unsupervised}, \cite{gidaris2018unsupervised}, and \cite{zhang2016colorful}, or instance-discrimination tasks, detailed in \cite{chen2020simple, wu2018unsupervised, oord2018representation, caron2018deep, asano2019self}. The underlying premise is that the features developed via this methodology are broadly applicable, as they can differentiate between a large number of instances or solve general pretext tasks. In videos, a similar approach is employed, with an increased focus on temporal aspects. Consequently, numerous pretext tasks have been proposed, including spatio-temporal puzzle solving \cite{kim2019self}, pace prediction \cite{wang2020self}, Odd-One-Out \cite{fernando2017self}, temporal order verification~\cite{misra2016shuffle,xu2019self}, speed prediction~\cite{chen2021rspnet,benaim2020speednet}, sorting sequences~\cite{lee2017unsupervised}, playback rate perception~\cite{yao2020video}, and ranking transformations~\cite{duan2022transrank}. For instance-discrimination, the methods such as~\cite{dave2022tclr,yao2021seco,qian2021spatiotemporal,wang2021enhancing,tao2020self,tubelet_fida} have been proposed, which set out to learn a representation that can separate between different instances, while ignoring the meaningless variances introduced by spatio-temporal data augmentations~\cite{dorkenwald2022scvrl,chen2020simple,he2020momentum,oord2018representation,thoker2023tubelet}. Unlike these methods, our approach does not depend on spatio-temporal augmentations. This enhances its generalizability and scalability, as it eliminates the need for dataset-specific augmentation design.\\

\noindent\textbf{Masked Input Modeling for Vision Transformers.} The idea of reconstructing masked input as a generative pretext task was first introduced by~\cite{pathak2016context}. MAE~\cite{he2022masked} showed that the features learned by this approach are very general when the backbone is replaced by vision transformers~\cite{dosovitskiy2020image}. Since then, lots of methods have been proposed that expand on the idea, such as~\cite{feichtenhofer2022masked, wang2022bevt, xie2022simmim, wei2022masked, NguyenICLR2024} in the image domain.  Recent research, including JEPA \cite{assran2023self}, has proposed the novel approach of mask feature prediction within the latent space, demonstrating improved performance on various image benchmarks, at the expense of instability challenges. VideoMAE~\cite{tong2022videomae} employs the same approach for the first time in video. Follow-up works have tried to improve VideoMAE by predicting low-level predefined targets instead of pixels~\cite{wang2023masked, wang2022bevt} or improve the temporal understanding of VideoMAE by either designing masks that mainly focus on the moving objects or explicitly adding motion information in the objective function~\cite{huang2023mgmae, mme_sun}. MME~\cite{mme_sun}, for instance, applies two major modifications to VideoMAE. First, following~\cite{feichtenhofer2022masked}, the decoder predicts HOG features of the masked area instead of the pixel values. Second, it extracts object motion trajectories using optical flow and predicts these trajectories in conjunction with the masked features using the decoder part. However, as motion trajectories become inaccurate in the presence of camera motion, the entire training dataset undergoes preprocessing to eliminate the effects of camera motion before training. By similar intention of learning better motion information, MGM~\cite{huang2023mgmae} and MGMAE use the hand-crafted motion vectors used by H.264 codec~\cite{richardson2002video} or optical flow modality to mask tubes representing moving objects more than other areas. Unlike such methods, our method is general and does not require any preprocessing or hand-crafted features. Instead, we propose to project spatio-temporal tubes in a deep feature space as the targets of the decoder to jointly learn high-level motion and appearance patterns.

\section{Methodology}
\begin{figure}[t]
    \centering
    \includegraphics[width=\textwidth]{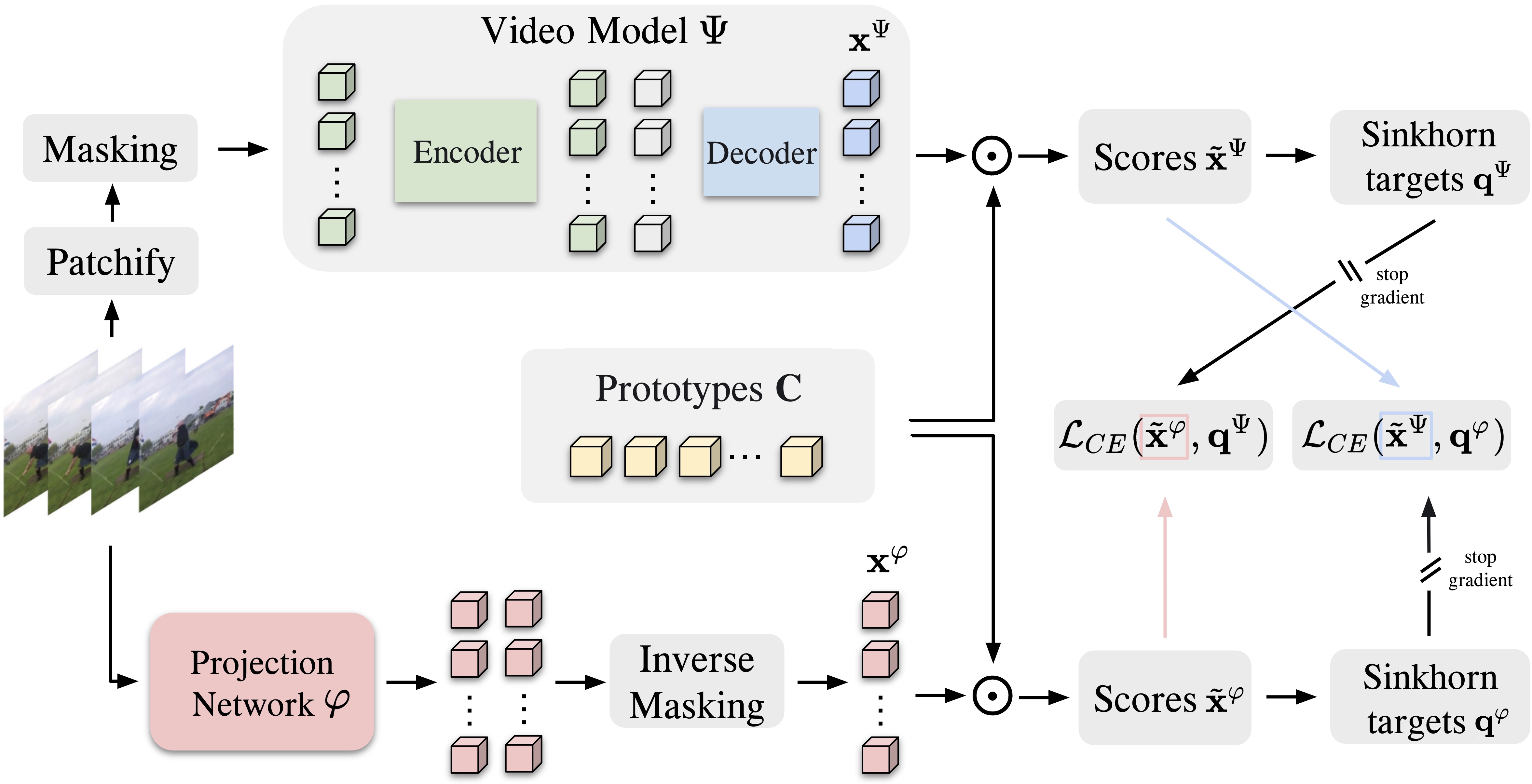}
    \caption{Overview of our proposed method \ours. A given video is embedded with the projection network \(\varphi\) leading to features \(\mathbf{x}^{\varphi}\). The video model \(\Psi\) predicts feature embeddings \(\mathbf{x}^\Psi\) of the masked space-time tubes. Both embeddings are projected onto the learnable prototypes representing cluster centroids. Cluster assignments are created with an adapted Sinkhorn algorithm enforcing equipartition across all prototypes. These pseudo-labels are then used as targets for the predictive task \(\mathcal{L}_{CE}\) with which the networks are optimized.}
    \label{fig:method}
\vspace{-0.3cm}
\end{figure}

In Sec.~\ref{sec:videomae}, we first review the common approach to masked video modeling, which focuses on the precise reconstruction of masked pixel values, leading to the extraction of low-level features. As a solution, we introduce a projection network \(\varphi\) in Sec.~\ref{sec:projection_network} that is jointly optimized with the video model and whose features are used as targets for the reconstruction task. However, this adjustment leads to a trivial solution. To circumvent this, we employ a cluster bottleneck strategy, as detailed in Section~\ref{sec:sinkhorn}. Features of similar space-time tubes are matched to the same cluster which infuses semantic meaning into our feature space. With these pseudo-labels, we can formulate a self-supervised prediction task with which we optimize the parameters of our networks. An overview of our method is depicted in Figure~\ref{fig:method}.

\subsection{Masked video modeling}
\label{sec:videomae}
In masked video modeling\cite{videomae, mme_sun, wang2023videomaev2, guided_masking_fan, cmae_v_lu}, the primary objective is the precise reconstruction of masked pixel values. Specifically, a video \(V\), composed of frames \(\{x_1, \dots, x_T\}\), is segmented into a set of space-time tubes \(\mathcal{T}\) with a significant proportion (approximately 90\%) being masked motivated by the inherent redundancy in video data. A video model \(\Psi\) with an asymmetric encoder-decoder architecture is trained to accurately reconstruct the pixels of the masked space-time tubes \(\mathcal{T}^M\) leveraging the unmasked tubes \( \mathcal{T}^{N} = \mathcal{T} \textbackslash \mathcal{T}^M \) 
and the positional information of the masked patches via
\begin{eqnarray}
    \mathcal{L}_{2} = \frac{1}{N} \sum_{i=1}^{N} \lVert \mathcal{T}_{i} - \Psi(\mathcal{T}^{N})_{i} \rVert_2^2,
\label{eq:reconstruction_loss}
\end{eqnarray}
with \(N\) the number of masked space-time tubes, \(\mathcal{T}_i\) denotes the pixels of the \(i\)-th space-time tube, and \(\Psi(\mathcal{T}^{N})_{i} \) represents the reconstructed pixels for this tube given the set of all non-masked tubes \(\mathcal{T}^{N}\). This loss function encourages strictly predicting the pixel values of the input video. Thus, mainly low-level features are captured in the trained video model \(\Psi\) as evidenced by low linear probing performance (see Tab. \ref{tab:linear-probing}). Yet, video reasoning necessitates an understanding of how scenes and objects interact with one another and transform across time. To achieve such a higher-level semantic understanding, we first exchange the target space in which the reconstruction loss is applied, to a deeper feature space.

\subsection{From pixel to feature reconstruction}
\label{sec:projection_network}
Rather than directly predicting pixel values, we change the target space to a feature space by introducing a projection network that can be jointly optimized. This enables the projection network to autonomously learn the target space, diverging from previous works that use predefined targets such as frame differences~\cite{yang2022self}, motion trajectories~\cite{mme_sun} or HOG features~\cite{feichtenhofer2022masked}.
The projection network \(\varphi\) embeds space-time tubes from a given video into features representing \(\mathbf{x}^{\varphi} = \varphi(\mathcal{T})  \) (see bottom stream in Fig.\ref{fig:method}).
We only select the features from the space-time tubes \(\mathcal{T}^M\) that are used for the masked prediction task in the video model and name this sampling process inverse masking in Fig. \ref{fig:method}.  
Thus, the target space has been replaced, and Eq. \ref{eq:reconstruction_loss} changes to 
\begin{eqnarray}
    \mathcal{L}^F_{2} = 
    \frac{1}{N} \sum_{i=1}^{N} \lVert \mathbf{x}^{\varphi}_{i} - \mathbf{x}^{\Psi}_{i} \rVert_2^2,
\label{eq:recon_feature_sapce}
\end{eqnarray}
with the prediction \(\mathbf{x}^{\Psi}  = \Psi(\mathcal{T}^{N})\). 
However, this leads to a trivial solution as both, the parameters of the projection network \(\varphi\) and the encoder-decoder video model \(\Psi\), are being jointly optimized, and can agree to map all space-time tubes to the same feature vector \( \varphi(\mathcal{T}^N) = \Psi(\mathcal{T}^N)\).

\subsection{\textsc{sigma}: Sinkhorn-Guided Masked Video Modeling}
\label{sec:sinkhorn}
To circumvent this trivial solution, we constrain the feature space to be partitioned into a limited number of clusters. This can be interpreted as a bottleneck with enforced high entropy. 
We accomplish this in an online manner by mapping the features \(\mathbf{x}\) of space-time tubes to a set of learnable prototypes \(\mathbf{C} = \{\mathbf{c}_1, \dots, \mathbf{c}_K\}\), representing cluster centroids. For that, we define a mapping \(\mathbf{Q}\in\mathbb{R}_+^{K\times B}\) of embeddings $\mathbf{X} = [\mathbf{x}_1, \dots, \mathbf{x}_B]$ to prototypes \( \mathbf{C} \).
An equipartition constraint is introduced on \(\mathbf{Q}\) to enforce that all prototypes are equally used, mitigating the trivial solution of all data points collapsing onto a single prototype. 
Due to the limited number of prototypes, similar and nearby space-time tubes are assigned to the similar prototypes, infusing semantic spatial and temporal meaning into the feature space.
This assignment can be formulated as an entropy-regularised optimal transport problem \cite{asano2020labelling} for all space-time tubes over a mini-batch \(\mathcal{B}\)
\begin{align}
      \min_{\mathbf{Q}} \langle \mathbf{Q}, - \log \mathbf{X} \rangle
  +  \frac{1}{\lambda} \operatorname{KL}(\mathbf{Q} \|rc^\top), \\
   \text{with}\,\, r = \frac{1}{K}\cdot\mathbbm{1},\quad c = \frac{1}{B}\cdot \mathbbm{1}.
\label{eq:asano}
\end{align}
Here $\lambda$ controls the entropy regularisation, $r,c$ are the marginals for the prototypes, and \(B=N_B \times N_\mathcal{T}\) being all samples \(N_B\) in the mini-batch \(\mathcal{B}\) for all space-time tubes \(N_\mathcal{T}
\). We rely on the Sinkhorn algorithm~\cite{cuturi2013sinkhorn} to solve Eq.~\ref{eq:asano}, which can be done extremely quickly on the GPU \cite{caron2020unsupervised, asano2020labelling} and yields soft pseudo-labels $\mathbf{q}$, such that $\text{argmax}(\mathbf{q}) = \mathbf{Q}$. 
With that, pseudo-labels \(\mathbf{q}^{\varphi}\) are generated for the embeddings \(\mathbf{x}^{\varphi}\) of the projection network. These pseudo-labels are then used as targets for the video model \(\Psi\) and vice versa.
Then the goal of \(\varphi\) and \(\Psi\) is to predict each other cluster assignments \(\mathbf{q} \) based on the projection of the features onto the shared prototypes \cite{caron2020unsupervised}. 
We formulate this predictive task between the pseudo-label \(\mathbf{q}\) and the scores \(\tilde{\mathbf{x}} = \mathbf{x}^\top \mathbf{c} \) using cross-entropy
\begin{align}
  \mathcal{L}_{CE}(\tilde{\mathbf{x}}, \mathbf{q}) = - \sum_{k} \mathbf{q}_{(k)} \log \mathbf{p}_{(k)} \\
  \mathbf{p}_{(k)} = \frac{ \exp \left ( \frac{1}{\tau} \tilde{\mathbf{x}}_k \right ) }{\sum_{k'} \exp \left ( \frac{1}{\tau} \tilde{\mathbf{x}}_{k'} \right ) }
  \label{eq:cross_entropy_loss}
\end{align}
where \(\tau\) represents the temperature parameter. We use softmax to define a probability distribution \(\mathbf{p_{(k)}}\) between \(\mathbf{x}\) and all \(K\) prototypes in \(\mathbf{C}\).
The loss function for all space-time tubes \(\mathcal{T}\) in a mini-batch \(\mathcal{B}\) can be written as a symmetric prediction task
\begin{eqnarray}
   \mathcal{L} =\frac{1}{B} \sum_{i=1}^{B} [ \mathcal{L}_{CE}(\tilde{\mathbf{x}}^{\varphi}_i, \mathbf{q}^{\Psi}_i) + \mathcal{L}_{CE}(\tilde{\mathbf{x}}^{\Psi}_i, \mathbf{q}^{\varphi}_i)],
\label{eq:symmetric_loss}
\end{eqnarray}
with \(\mathcal{L}_{CE}(\tilde{\mathbf{x}}^{\varphi}, \mathbf{q}^{\Psi})\) the loss for predicting the pseudo-labels based on the feature of the video model \(\Psi\) and \(\mathcal{L}_{CE}(\tilde{\mathbf{x}}^{\Psi}, \mathbf{q}^{\varphi})\) the one from the projection network \(\varphi\). By using feature pseudo-labels  as targets for our symmetric loss, we solve the shortcomings of low-level predefined targets in Eq. \ref{eq:reconstruction_loss} and circumvent the trivial solution for deep targets outlined in Eq. \ref{eq:recon_feature_sapce}.
In contrast to prior works\cite{qian2021spatiotemporal, bardes2023mcjepa, svt, tubelet_fida, patrick2020multi, dorkenwald2022scvrl} that rely on handcrafted augmentations to generate different views for their loss objective, our loss Eq.~\ref{eq:symmetric_loss} circumvents that by instead predicting deep feature targets of masked space-time tubes. The joint space defined by the Sinkhorn clustering also mitigates the need for a momentum encoder as required by previous works \cite{bardes2023mcjepa, qian2021spatiotemporal, dorkenwald2022scvrl} and
thus allows us to vary the architecture choice for our projection network \(\varphi\), as we will demonstrate in our experiments.

\section{Experiments}
\label{sec:experiments}
We evaluate our method on a total of ten different datasets and across three common benchmark settings. 
In Sec.~\ref{common} we first compare our approach against state-of-the-art video models in a linear probing (frozen backbone) and the standard full finetuning setting.
Then, in Sec.~\ref{segment} we benchmark the semantic spatial and temporal understanding by reporting unsupervised semantic segmentation performance and visualizing the segmentation masks. 
In Sec.~\ref{severe} we evaluate our approach on the SEVERE benchmark~\cite{thoker2022severe, tubelet_fida} specifically designed to analyze the generalization performance of video models. Lastly, we ablate parts of our approach to give more insights in Sec.~\ref{ablate}. 

\paragraph{Implementation details.} 
As a projection network \( \varphi\) we use a simple 3-layer Multi-Layer Perceptron (MLP) consisting of a linear layer with 1024 neurons followed by a GELU activation function and a final linear layer mapping to the feature dimension of \(256\) for the predictive task. This MLP embeds a given space-time tube and is trained jointly with our video model \( \phi\). This variant is abbreviated as \ours w/ MLP. Moreover, our approach allows us to leverage pretrained models such as DINO~\cite{caron2021emerging} which is kept frozen while training the video model (\ours w/ DINO). We pretrain these variants on: \textbf{{Something-Something V2 (SSv2)}}~\cite{goyal2017something} 
and \textbf{Kinetics-400 (K400)}. SSV2 contains 220K videos with 174 action classes and is considered motion-heavy because of its focus on motion and directional aspects inherent to the actions. K400 is acknowledged as the benchmark for video recognition evaluations, encompassing 240K Internet-sourced videos classified into 400 action categories. We follow previous works for pretraining and use a temporal stride \(\tau\) of 2 for SSv2 and 4 for K400. We sample a clip of 16 frames from the raw video at a resolution of \( 224 \times 224 \) pixels. We use space-time tube embeddings like in VideoMAE\cite{videomae} where each cube of size \( 2 \times 16 \times 16\) is treated as one token embedding and are extracted with a 3D convolution layer. We also use the commonly high masking ratio of \(90\%\) with which tubes are randomly masked. To be comparable with previous works and use the same configurations for the small and base ViT backbones. We follow \cite{videomae} to use AdamW\cite{adamw} optimizer with a base learning rate \( 1.5e^{-4} \), weight decay \( 0.05\), \( \beta= [0.9, 0.95]\), and cosine learning rate decay. For downstream tasks, we employ average pooling upon all output tokens before a final layer is added to map from the feature dimension to the number of classes used for classification.
We follow the full finetuning setup from VideoMAE\cite{videomae} and list all hyperparameters for pretraining and finetuning in the supplemental material. The source code of this project will be released upon acceptance.

\subsection{Benchmark I: Comparisons for linear and full finetuning} \label{common}
 
We conduct the linear and full finetuning experiments on four commonly used datasets including \textbf{SSv2}, \textbf{K400}, \textbf{UCF-101}~\cite{soomro2012ucf101} and  \textbf{{HMDB-51}}~\cite{kuehne2011hmdb}. We refer to the supplementary material for details on the datasets. 
In addition, we also show the effectiveness of video models on image datasets such as \textbf{ImageNet-1K} (IN-1K)~\cite{imagenet}, and \textbf{CIFAR-100} (C-100) \cite{cifar}.

\begin{table}[t]
    \centering
    \caption{\textbf{Benchmark I: Frozen evaluation of masked video modeling methods.} A linear layer on top of the frozen ViT-B backbone is optimized. The ViT-B backbones are pretrained on Kinetics-400 (K400). We evaluated the official publicly released model from the corresponding method. \ours consistently outperforms previous masked video modeling works across all video and image datasets, often considerably.}
    \setlength{\tabcolsep}{0.5em}
    \begin{tabular}{lccccccc} \toprule & SSv2  & K400 &  UCF & HMDB & IN-1K & {C-100} \\
         \midrule
         MVD \cite{mvd_wang}&12.2 & 18.7 & 49.1 & 28.6& 20.0 & 34.1\\
         MME \cite{mme_sun}& 16.6 & 19.1    & 56.0 & 37.1 & 15.8 & 34.2\\
         VideoMAE \cite{videomae} & 17.5 &20.7 & 58.6 & 37.7 & 20.2 & 40.4 \\
         MGMAE \cite{mgmae_huang} & 16.8 & 24.9 & 64.4  & 41.3 & 20.4 & 44.1 \\
         \rowcolor{lightcyan}
          \ours w/ MLP &  \underline{19.9} &  \underline{30.7} &  \underline{73.8} &  \underline{45.0} & \underline{24.1} & \underline{46.2} \\ %
          \rowcolor{lightcyan}
         \ours w/ DINO  & \textbf{20.8} & \textbf{47.5} & \textbf{80.7} & \textbf{52.3} & \textbf{45.0} & \textbf{66.7} \\
        \bottomrule
    \end{tabular}

    \label{tab:linear-probing}
\vspace{-0.3cm}
\end{table}

\paragraph{Frozen evaluation setup.} We analyze the downstream abilities of our approach and compare it against state-of-the-art methods using a frozen evaluation setting. The underlying ViT-B backbone pretrained on K400 remains frozen while a linear layer on top is optimized for the specific target dataset. This frozen setup better evaluates the pretrained models compared to full finetuning where all the learned parameters are overwritten. We compare against the recent masked video modeling works MGM\cite{guided_masking_fan}, MME\cite{mme_sun}, MVD\cite{mvd_wang}, and VideoMAE\cite{videomae}. We used the official publicly released pretrained models from those methods and used the commonly used 800 epochs checkpoint except for MVD which pretrain for 1600 epochs and distill for 400 epochs. %

\paragraph{Frozen evaluation results.}
From the results in Tab.~\ref{tab:linear-probing} we observe a big improvement in performance with our method against all state-of-the-art methods while using only a simple MLP as a projection network leading to an average improvement of 4.5\( \%\) across all datasets and domains. Furthermore, an even bigger improvement is observed when we train our method using DINO as a projection network with an average improvement of 16.8\( \%\) across all datasets. Surprisingly, we observe that MVD, which relies on predefined MAE feature targets, achieves the lowest performance, showcasing that the ability of our method to jointly learn a more semantic target space is crucial. 
Our method outperforms VideoMAE and MGM, which use pixel values as reconstruction targets, highlighting the efficacy of our Sinkhorn-guided approach that predicts feature cluster assignments.
In summary, these findings confirm that our approach leads to more spatial and temporal semantic features.

\begin{table*}[t]
\centering
\caption{\textbf{Benchmark I: Comparison for full finetuning on Something-Something V2} (SSv2). The top part compromises supervised methods while the remaining methods are pretrained in a self-supervised manner. The middle section evaluates models trained on Kinetics 400 (K400) data for pretraining whereas the bottom part mainly used SSv2 data. We compare against previous methods pretrained on the ViT-Base backbone for 800 epochs. A full table with all previous works using also different pretraining setups is provided in the supplemental. M. Guid. denotes motion guidance such as optical flow used e.g. reconstructing targets or masking. Our method achieves state-of-the-art performance on SSv2 w/ and w/o motion guidance.}
\resizebox{0.9\textwidth}{!}{%
\begin{tabular}{ll cccccccc}
\toprule
&Method & M. Guid. & Backbone & Epochs & Extra data & Frames  & Params & Top-1 \\
\midrule
&\textbf{\textit{supervised baselines}} & & & & & & & & \\
&\quad SlowFast\cite{Feichtenhofer2018SlowFastNF} & -- & ResNet101 & -- & K400 & 8+32  & 53 & 63.1 \\
&\quad MViTv1~\cite{mvit_v1_fan}  &--& MViTv1-B & -- & K400 & 64 & 37 & 67.7 \\
&\quad TimeSformer~\cite{Bertasius2021IsSA} & --& ViT-B & -- & IN-21K & 8 & 121 & 59.5 \\
&\quad VideoSwin~\cite{videoswin_liu} & --&Swin-B & -- & IN-21K & 32 & 88 & 69.6 \\
\midrule
&\textbf{\textit{self-supervised}} & & & & & & & \\
\parbox[t]{3mm}{\multirow{5}{*}{\rotatebox[origin=c]{90}{\textbf{K400 pretraining}}}} &\quad \cellcolor{lightcyan}\textsc{sigma-dino} (ours) & \cellcolor{lightcyan}$\times $& \cellcolor{lightcyan}ViT-S  & \cellcolor{lightcyan}800& \cellcolor{lightcyan}IN-1K+K400 &\cellcolor{lightcyan}16& \cellcolor{lightcyan}22 & \cellcolor{lightcyan} 68.7\\
\cmidrule{3-9}
&\quad OmniMAE~\cite{girdhar2023omnimae} & $\times$&  ViT-B & 800 & IN-1K+K400 &16&  86 & 69.0 \\
&\quad VideoMAE~\cite{videomae} & $\times$& ViT-B  & 800 & K400 & 16 & 87 & 68.5 \\
&\quad MME~\cite{mme_sun}& $\checkmark$ & ViT-B  & 800 & K400 & 16 & 87 & \underline{70.5} \\
&\quad \cellcolor{lightcyan}\textsc{sigma-mlp} (ours) &  \cellcolor{lightcyan}$\times$&  \cellcolor{lightcyan}ViT-B  & \cellcolor{lightcyan}800& \cellcolor{lightcyan}K400&\cellcolor{lightcyan}16& \cellcolor{lightcyan}87 & \cellcolor{lightcyan}69.8 \\
&\quad \cellcolor{lightcyan}\textsc{sigma-dino} (ours)&  \cellcolor{lightcyan}$\times$&\cellcolor{lightcyan}ViT-B  & \cellcolor{lightcyan}800& \cellcolor{lightcyan}IN-1K+K400 &\cellcolor{lightcyan}16& \cellcolor{lightcyan}87 & \cellcolor{lightcyan}\textbf{71.1} \\
\cmidrule{2-9}
&\quad VideoMAE~\cite{videomae}& $\times$& ViT-S  & 2400 & -- & 16  & 22 & 66.8 \\
&\quad \cellcolor{lightcyan}\textsc{sigma-dino} (ours) & \cellcolor{lightcyan}$\times$& \cellcolor{lightcyan}ViT-S  & \cellcolor{lightcyan}2400& \cellcolor{lightcyan}IN-1K &\cellcolor{lightcyan}16 & \cellcolor{lightcyan}22 & \cellcolor{lightcyan}\textbf{68.6} \\
\cmidrule{3-9}
\parbox[t]{3mm}{\multirow{8}{*}{\rotatebox[origin=c]{90}{\textbf{SSv2 pretraining}}}} &\quad OmniMAE~\cite{girdhar2023omnimae} & $\times$& ViT-B & 800 & IN-1K & 16& 86 & 69.5 \\
&\quad VideoMAE~\cite{videomae}& $\times$& ViT-B & 800 &-- & 16  & 87 & 69.6\\
 &\quad CMAE-V~\cite{cmae_v_lu} & $\times$& ViT-B  & 800 &  -- & 16 & 87 & 69.7 \\
&\quad \cellcolor{lightcyan}\textsc{sigma-mlp} (ours) & \cellcolor{lightcyan}$\times$ & \cellcolor{lightcyan}ViT-B  & \cellcolor{lightcyan}800& \cellcolor{lightcyan}-- &\cellcolor{lightcyan}16 & \cellcolor{lightcyan}87 &\cellcolor{lightcyan}\textbf{70.4} \\
&\quad \cellcolor{lightcyan}\textsc{sigma-dino} (ours) & \cellcolor{lightcyan}$\times$ & \cellcolor{lightcyan}ViT-B  & \cellcolor{lightcyan}800& \cellcolor{lightcyan}IN-1K &\cellcolor{lightcyan}16 & \cellcolor{lightcyan}87 & \cellcolor{lightcyan}\textbf{70.9} \\
\cmidrule{3-9}
&\quad MME~\cite{mme_sun}& $\checkmark$& ViT-B  & 800 & -- & 16& 87 & 70.0 \\
&\quad MGM~\cite{guided_masking_fan} & $\checkmark$ & ViT-B   & 800 & --  &16& 87 &  70.6 \\
&\quad 
MGMAE~\cite{mgmae_huang} & $\checkmark$ & ViT-B   & 800 & --  &16& 87 &  {71.0} \\
&\quad \cellcolor{lightcyan}\textsc{sigma-mlp} (ours) & \cellcolor{lightcyan}$\checkmark$ & \cellcolor{lightcyan}ViT-B  & \cellcolor{lightcyan}800& \cellcolor{lightcyan}-- &\cellcolor{lightcyan}16 & \cellcolor{lightcyan}87 & \cellcolor{lightcyan}\textbf{71.2} \\
&\quad \cellcolor{lightcyan}\textsc{sigma-dino} (ours) & \cellcolor{lightcyan}$\checkmark$ & \cellcolor{lightcyan}ViT-B  & \cellcolor{lightcyan}800& \cellcolor{lightcyan}IN-1K &\cellcolor{lightcyan}16 & \cellcolor{lightcyan}87 & \cellcolor{lightcyan}\textbf{71.2} \\
\bottomrule
\end{tabular}%
}

\label{tab:ssv2}
\vspace{-0.3cm}
\end{table*}

\begin{table*}[t]
\centering
\caption{\textbf{Benchmark I: Comparison for full finetuning on Kinetics 400} (K400). We compare against previous methods for pretraining the ViT-Base backbone for 800 epochs on K400 and subsequently, fully finetuning the backbone with the K400 labels. A full table with all previous methods using also different setups is provided in the supplemental. M. Guid. denotes motion guidance such as optical flow used e.g. reconstructing targets or masking. Our method achieves state-of-the-art performance on K400.}
\resizebox{0.88\textwidth}{!}{%
\begin{tabular}{ll cccccccc}
\toprule
&Method & M. Guid. & Backbone & Epochs & Extra data & Frames & Params & Top-1 \\
\midrule
&\textbf{\textit{supervised}} & & & & & & & & \\
&\quad SlowFast~\cite{Feichtenhofer2018SlowFastNF}  & $\times$ & ResNet101 & -- & -- & 16+64 & 60 & 79.8  \\
&\quad MViTv1~\cite{mvit_v1_fan} & $\times$ & MViTv1-B & -- & -- & 32  & 37 & 80.2 \\
&\quad TimeSformer~\cite{Bertasius2021IsSA} & $\times$ & ViT-B & -- & IN-21K & 96 & 430 & 80.7\\
&\quad VideoSwin~\cite{videoswin_liu} & $\times$ & Swin-L & -- & IN-21K & 32& 197 & 83.1 \\
\midrule
&\textbf{\textit{self-supervised}} & &  & & & & & & \\
&\quad VideoMAE~\cite{videomae} & $\times$ & ViT-S  & 1600 &-- & 16 & 87 & 79.0 \\
\parbox[t]{3mm}{\multirow{5}{*}{\rotatebox[origin=c]{90}{\textbf{K400 pretraining}}}}&\quad \cellcolor{lightcyan}\textsc{sigma-dino} (ours) & \cellcolor{lightcyan}$\times$ & \cellcolor{lightcyan}ViT-S  & \cellcolor{lightcyan}800& \cellcolor{lightcyan}IN-1K &\cellcolor{lightcyan}16 &\cellcolor{lightcyan} 87 & \cellcolor{lightcyan} \textbf{79.4} \\
\cmidrule{3-9}
&\quad VideoMAE~\cite{videomae} & $\times$ & ViT-B  & 800 & -- & 16  & 87 &  80.0 \\
&\quad OmniMAE~\cite{girdhar2023omnimae} & $\times$ & ViT-B & 800 & IN-1K & 16& 87 & 80.8 \\
&\quad CMAE-V~\cite{cmae_v_lu} & $\times$ & ViT-B  & 800 &  -- & 16  & 87 & 80.2 \\
&\quad MGM~\cite{guided_masking_fan}& $\checkmark$ & ViT-B   & 800 & -- & 16& 87 &  80.8 \\
&\quad MGMAE~\cite{mgmae_huang}& $\checkmark$ & ViT-B   & 800 & -- & 16& 87 &  \underline{81.2} \\
&\quad \cellcolor{lightcyan}\textsc{sigma-mlp}  (ours)& \cellcolor{lightcyan}$\times$ & 
\cellcolor{lightcyan}ViT-B  & \cellcolor{lightcyan}800& \cellcolor{lightcyan}-- &\cellcolor{lightcyan}16& \cellcolor{lightcyan}87 & \cellcolor{lightcyan}80.2 \\
&\quad \cellcolor{lightcyan}\textsc{sigma-dino} (ours)
 & \cellcolor{lightcyan}$\times$ &\cellcolor{lightcyan}ViT-B  &\cellcolor{lightcyan}800
&\cellcolor{lightcyan} IN-1K &\cellcolor{lightcyan}16 & \cellcolor{lightcyan}87 & \cellcolor{lightcyan}\textbf{81.6}\\
\bottomrule
\end{tabular}%
}

\label{tab:k400}
\end{table*}

\paragraph{Full finetuning setup.} In this experiment, we evaluate the downstream action classification abilities of our method in a full finetuning setup for SSv2 and K400. We use the ViT-S and ViT-B backbones and pretrain for 800 epochs on the respective datasets following previous works~\cite{videomae, guided_masking_fan, cmae_v_lu}. 
For clarity, we compare against methods with this commonly used pretraining setup and report a table featuring all previous works in the supplemental material.

\paragraph{Full finetuning results.} We report results for SSv2 in Tab.~\ref{tab:ssv2} and for K400 in Tab.~\ref{tab:k400}. Given the nature of full finetuning in which all learned parameters of the network are optimized the gap between methods is smaller. Still, we observe a substantial improvement of on average \(\sim\!1 \%\) across all datasets and backbones from our method using a simple MLP as a projection network compared to VideoMAE~\cite{videomae}. This confirms the importance of utilizing a learnable target reconstruction space rather than simply regressing pixel values. \ours can also incorporate motion guidance via advanced masking following MGMAE~\cite{mgmae_huang}, further improving the results on SSv2, especially for the MLP variant.
Our method with DINO as a projection network achieves state-of-the-art performance on SSv2 and K400. 
We even outperform the VideoMAE ViT-B model with our ViT-S K400 pretrained model on SSv2.

\begin{figure}[t]
    \centering
    \includegraphics[width=\textwidth]{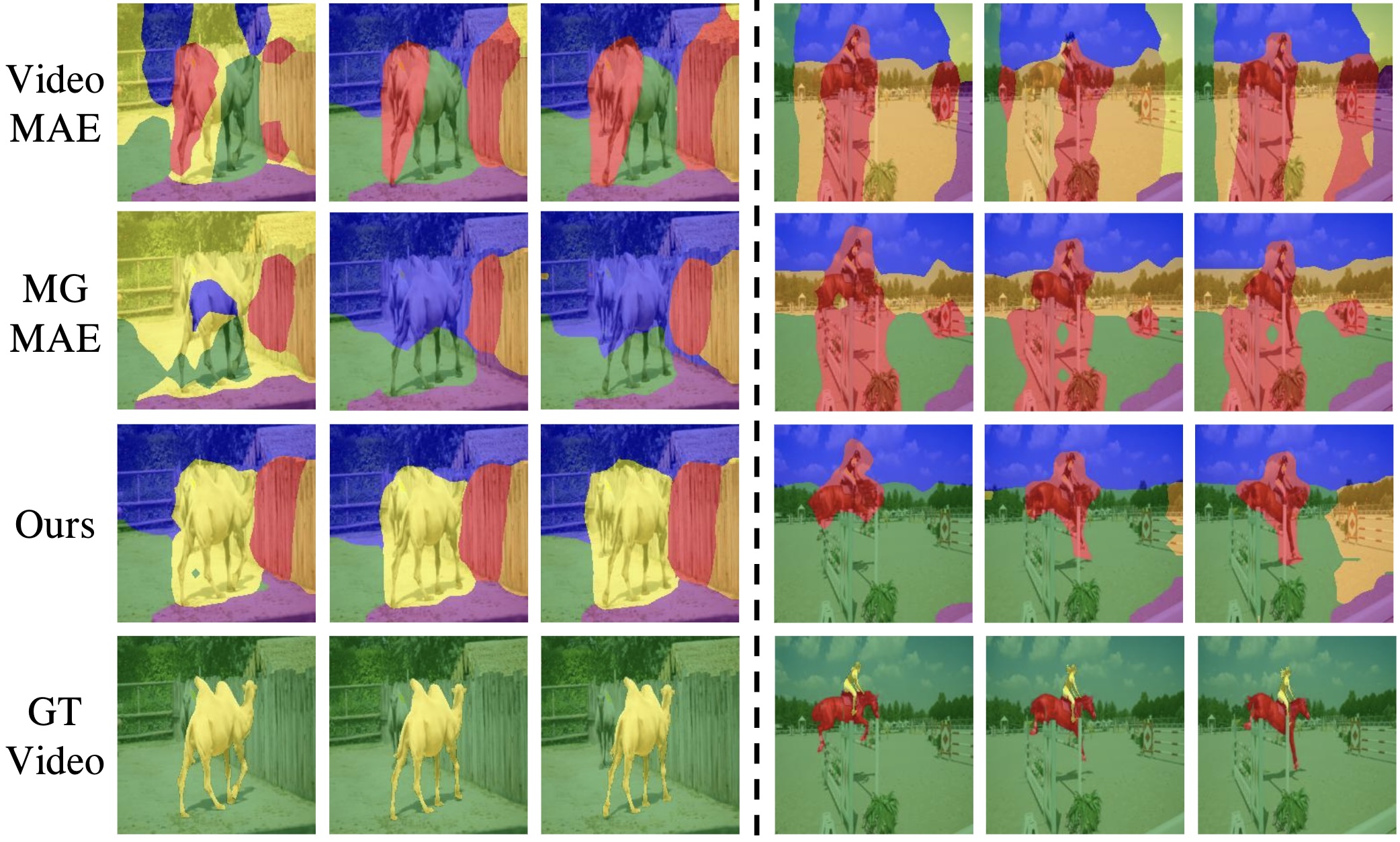}
    \caption{\textbf{Benchmark II: Unsupervised video object segmentation results on DAVIS.} We visualize the abilities of masked video modeling methods to produce temporally consistent semantic segmentation masks. %
    \textsc{Sigma} provides more coherent and consistent object cluster maps compared to other methods. This shows that our learned features have better temporal and spatial understanding.}
    \label{fig:object-segmentation}
\vspace{-0.7cm}
\end{figure}

\subsection{Benchmark II: Unsupervised video object segmentation} \label{segment}

\paragraph{Setup.} To assess the temporal and spatial semantics learned by our proposed method, we employ the unsupervised video object segmentation benchmark introduced by~\cite{salehi2023time}. This benchmark evaluates the pretrained video encoder's ability to generate temporally consistent object segmentation maps through its space-time features. In this regard, it deviates from the other reported evaluation settings where dense space-time features are pooled for a global clip representation. Instead, space-time features are grouped into clusters via k-means, based on a predefined number of clusters denoted by the $K$ parameter. These clusters are subsequently matched against ground truth object segmentation maps using the Hungarian algorithm~\cite{kuhn1955hungarian}, and their overlap (mIoU) is calculated to determine the unsupervised object segmentation score. This procedure is called clustering when $K$ matches the count of objects in the ground truth video, and overclustering when $K$ exceeds this count. For both procedures, we report the segmentation performance in terms of mIoU for two datasets: \textbf{DAVIS}~\cite{pont20172017} and \textbf{YTVOS}~\cite{xu2018youtube}.
Please see the supplementary materials for details on the datasets.

\paragraph{Results.} As shown in Tab.~\ref{table:clustering_overclustering}, our method with a simple MLP as a projection layer outperforms almost all the competitors by 1.5\% on average. \textsc{sigma} with DINO as a projection network works even better and outperforms all previous works by a large margin across all metrics by approximately 3\% on average using the ViT-B backbone and by approximately 4.5\% with ViT-S. These findings indicate that our model produces more semantic space-time representations. 
Moreover, although MGM~\cite{huang2023mgmae} concentrates on developing masks for moving objects to enhance feature representation, our method still achieves a 2\% to 5\% improvement without changing the commonly used random tube masking. This result emphasizes the efficacy of \textsc{sigma} over methods based on pixel-level reconstruction for dense downstream tasks. 
In Fig.~\ref{fig:object-segmentation} we qualitatively compare the cluster maps produced using our encoder's features compared to VideoMAE~\cite{tong2022videomae} and MGM~\cite{huang2023mgmae} on DAVIS. For that, each clip is passed to the encoder and its tube representations are extracted. Then, the extracted features are clustered by k-means with a $K$ higher than object counts specified by the ground truth. The reason is, for DAVIS the number of ground truth objects is usually smaller than the real number of objects. Hence, overclustering is a more realistic setting. Finally, the extracted cluster maps are resized to match the input size and overlayed on the input. 
\textsc{sigma} with DINO as a projection network produces accurate segmentations for each video clip, indicating a better understanding of the object-level concepts. This can be seen for example in the left visualization in Fig. \ref{fig:object-segmentation} where the camel is well segmented, compared to VideoMAE and MGM. Additionally, our learned features exhibit better consistency over time than the temporal flickering observed in VideoMAE and MGM. These results indicate that our method leads to more robust video representations with object temporal consistency.

\begin{table*}[t]
\centering
{
\caption{\textbf{Benchmark II: Unsupervised video object segmentation.} We follow the evaluation protocol from \cite{salehi2023time} and report mIoU for clustering and overclustering. \ours consistently achieves better results compared to other methods across different backbones and datasets. This shows that our method learns better semantic and temporal consistent features. %
}
\label{table:clustering_overclustering}
\begin{tabular}{l cc cccc}
\toprule
& & \multicolumn{2}{c}{\textbf{Clustering}} & \multicolumn{2}{c}{\textbf{Overclustering}} \\
\cmidrule{3-4} \cmidrule{5-6}
 \textbf{Method} & \textbf{Backbone} & \multicolumn{1}{c}{YTVOS} & \multicolumn{1}{c}{DAVIS} & \multicolumn{1}{c}{YTVOS} & \multicolumn{1}{c}{DAVIS} \\
\midrule
VideoMAE \cite{videomae} & ViT-S & 33.7 & 28.3 & 53.9 & 52.8 \\
MVD \cite{mvd_wang}& ViT-S & 32.9 & 27.0 & 50.3 & 43.3 \\
\rowcolor{lightcyan}
\textsc{sigma-dino} (ours) & ViT-S & \textbf{38.2} & \textbf{30.6} & \textbf{60.0} & \textbf{56.8} \\
\midrule
VideoMAE \cite{videomae}& ViT-B & 34.1 & 27.2 & 49.7 & 50.9 \\
MVD \cite{mvd_wang}& ViT-B & 33.1 & 29.5 & 52.9 & 45.2 \\
MGMAE \cite{mgmae_huang} & ViT-B & 34.5 & 28.2 & 55.4 & \underline{54.8} \\
\rowcolor{lightcyan}
\textsc{sigma-mlp} (ours) & ViT-B & \underline{35.9} & \underline{30.2} & \underline{57.4} & 52.9 \\
\rowcolor{lightcyan}
\textsc{sigma-dino} (ours) & ViT-B & \textbf{36.4} & \textbf{30.6} & \textbf{60.4} & \textbf{56.5} \\
\bottomrule
\end{tabular}}

\vspace{-0.4cm}
\end{table*}

\subsection{Benchmark III: SEVERE generalization} \label{severe}
\paragraph{Setup.} 
To investigate the generalization capability of self-supervised video representations learned by our model we evaluate it on the challenging SEVERE benchmark introduced in \cite{thoker2022severe} and extended by~\cite{thoker2023tubelet}. This benchmark comprises eight experiments focused on four downstream generalization factors: \textit{domain shift}, \textit{sample efficiency}, \textit{action granularity}, and \textit{task shift}. Domain shift is assessed using \textbf{Something-Something v2} and \textbf{FineGym(Gym99)}~\cite{shao2020finegym}, both differing in domain relative to pretraining dataset Kinetics-400. Sample efficiency is tested through low-shot action recognition on \textbf{UCF101}  and \textbf{FineGym}, with only 1,000 training samples available for finetuning. Action granularity is explored by evaluating semantically similar actions using subsets \textbf{FX-S1} and \textbf{UB-S1} from FineGym, where action classes pertain to the same element of a gymnastic routine (e.g., FX-S1 representing types of jumps). Task shift assesses performance beyond single-label action recognition, involving temporal repetition counting on UCFRep~\cite{ucfrep-zhang2020context}, a subset of UCF101, and multi-label action recognition on \textbf{Charades}~\cite{sigurdsson2016hollywood}. Detailed experimental setups for each subset are provided in the supplementary. %

\paragraph{Results.}
 We follow the same setup as in the original SEVERE benchmark~\cite{thoker2022severe, thoker2023tubelet}. In particular, we compare with VideoMAE~\cite{videomae} and SVT~\cite{svt} as reported in \cite{thoker2023tubelet} and extend the benchmark with \textsc{sigma}, MVD\cite{mvd_wang}, and MGM\cite{guided_masking_fan}. All models are pretrained on Kinetics-400 using the ViT-B architecture. Results are shown in Table~\ref{tab:severe-performance}. \noindent 
\textbf{Domain Shift.} Our approach \textsc{sigma} (both with MLP and DINO) outperforms all prior methods on Gym99, \eg +7.2\% over MVD and +3.1\% over VideoMAE, demonstrating that the representation learned by our model is robust to domain shift in the downstream dataset. This is also demonstrated by strong SSv2 fine-tuning performance with K400 pretraining, as shown in Table~\ref{tab:ssv2}. %
\noindent \textbf{Sample Efficiency.}
For sample efficiency, we achieve a considerable gain over all prior works on Gym ($10^{3}$), \eg, +2.1\% over VideoMAE~\cite{videomae} and +4.0\%  over MGMAE~\cite{huang2023mgmae} showing the effective adaptability of our model to low shot action recognition. %
\noindent \textbf{Action granularity.} For fine-grained actions in FX-S1 and UB-S1, our method achieves the best performance as well, with a considerable improvement over other models, \eg, +12.3\% and +14.6\% over VideoMAE. These results demonstrate that the video representation learned by our method is better suited to fine-grained actions  than existing self-supervised methods.
\noindent \textbf{Task Shift.} For the task shift to repetition counting our method achieves the best performance. For multi-label action recognition on Charades, our approach lags behind SVT which employs a contrastive learning paradigm, however, we outperform all masked video modeling methods.
 \textbf{Overall SEVERE Performance.} 
Finally, we compare the mean across all generalizability factors. Our method has the best mean performance when using DINO as the projection network and achieves the second-best performance when employing MLP as the projection network. We conclude our method improves the generalizability of video self-supervised representations across these four downstream factors compared to all previous masked video modeling methods. 

\begin{table*}[t!]
    \centering
        \caption{\textbf{Benchmark III: SEVERE Generalization \cite{thoker2022severe}.}  We evaluate masked video modeling methods for generalizability in domain shift, sample efficiency, action granularity, and task shift following \cite{thoker2022severe}. SIGMA achieves strong generalization performance outperforming prior works across all configurations. We use the original severe codebase\cite{thoker2022severe,tubelet_fida} to evaluate the publicly available models for all the methods.}
        
    \setlength{\tabcolsep}{5pt}
    \resizebox{\linewidth}{!}{
    \begin{tabular}{l ccccc cccc c}
    \toprule
    & \multicolumn{2}{c}{\textbf{Domains}} & \multicolumn{2}{c}{\textbf{Samples ($10^{3}$) }} & \multicolumn{2}{c}{\textbf{Actions}} & \multicolumn{2}{c}{\textbf{Tasks}} & \textbf{Mean} \\
    \cmidrule(lr){2-3} \cmidrule(lr){4-5} \cmidrule(lr){6-7} \cmidrule(lr){8-9} \cmidrule(lr){10-10}
    & SSv2 &Gym99 & UCF & Gym & FX-S1 & UB-S1 & UCF-RC$\downarrow$ & Charades & \\
    \midrule
    SVT~\cite{svt}  & 59.5 & 62.3 & \underline{83.9} & 18.5 & 35.4 & 55.1 & 0.421 & \textbf{35.5} & 51.0 \\
    MVD~\cite{mvd_wang}& \underline{70.0} & 82.5 & 66.7 & 17.5 & 31.3 & 50.5 & 0.184 & 16.1 & 52.1 \\
    VideoMAE~\cite{tong2022videomae}  & 68.6 & 86.6 & 74.6 & 25.9 & 42.8 & 65.3 & \underline{0.172} & 12.6 & 57.4 \\
    MGMAE~\cite{mgmae_huang} & 68.9 & 87.2 & 77.2 & 24.0 & 33.7 & 79.5 & 0.181 & 17.9 & 58.8 \\
    \textsc{sigma-mlp} (ours)& {69.8}  & \underline{87.4} & 80.2 & \underline{26.8} & \underline{46.0} & \underline{79.7} & 0.178 & 20.1 & \underline{61.5} \\
    \textsc{sigma-dino} (ours) &\textbf{70.9} & \textbf{89.7} & \textbf{84.1} & \textbf{28.0} & \textbf{55.1} & \textbf{79.9} & \textbf{0.169} & \underline{23.3} & \textbf{64.3} \\
    \bottomrule
    \end{tabular}
    }

    \label{tab:severe-performance}
\vspace{-0.3cm}
\end{table*}

\subsection{Ablations} \label{ablate}
In this section, we ablate the hyperparameters of our proposed method \ours. For that, we use the Vit-S backbone and train and evaluate (full finetuning) on mini SSv2, a subset with 50\(\%\) of the data.  

\paragraph{Number of prototypes $\mathbf{C}$.} In Tab.~\ref{tab:ablation1} we report the effect of changing the number of prototypes on the performance of \ours. As the number of prototypes increases, the performance also increases and peaks at the moderate number of 4k. Also, the proposed method consistently improves upon VideoMAE~\cite{tong2022videomae} regardless of the number of prototypes. 

\paragraph{Projection network architecture.} 
We show the effect of changing MLP from a base architecture to a shallower, deeper, and wider version and report the performance in Tab.~\ref{tab:ablation_loss_combined} (left). Shallower architecture in the context of projection networks refers to reducing the number of layers. This results in a network with only 2 layers, each containing 1024 neurons. On the other hand, deeper or wider architectures involve adding more layers (4 instead of 3) or increasing the number of neurons in the same number of layers (2048 instead of 1024). The best performance is achieved when our base MLP projection network is used. Also, the slight differences in performance across various architectures highlight the robustness of the proposed method.\\

\paragraph{Loss function.} Tab.~\ref{tab:ablation_loss_combined} (right) presents an ablation study of various training objective functions analyzed with our framework. The L2 loss, when combined with deep features as targets, yields significantly lower results in comparison to our clustering-based approach, due to finding trivial shortcut solutions, as discussed in Sec.\ref{sec:projection_network}.

\begin{table*}[t]
\centering
\caption{\textbf{Ablating the number of prototypes} for SIGMA with MLP (left) and with DINO (right).
A moderate number of prototypes obtain the best performance. 
}   %
\label{tab:ablation1}

\begin{subtable}{.4\linewidth}
\centering
\begin{tabular}{ccc}
\toprule
Method & $\#$ Prototypes & Top-1 \\
\midrule
 VideoMAE & -- & 56.9 \\
\textsc{sigma-mlp} & 1000 & 59.7 \\
\textsc{sigma-mlp} & 2000 & 59.7 \\
\textsc{sigma-mlp}& 3000 & 59.9 \\
\cellcolor{lightcyan}\textsc{sigma-mlp} & \cellcolor{lightcyan}4000 & \cellcolor{lightcyan}\textbf{60.3} \\
\textsc{sigma-mlp} & 6000 & 60.1 \\
\bottomrule
\end{tabular}
\end{subtable}%
\begin{subtable}{.4\linewidth}
\centering
\begin{tabular}{ccc}
\toprule
Method & $\#$ Prototypes  & Top-1 \\
\midrule
 VideoMAE & -- & 56.9 \\
\textsc{sigma-dino} & 1000 & 62.8 \\
\textsc{sigma-dino} & 2000 & 62.8 \\
\rowcolor{lightcyan}
\textsc{sigma-dino}& 3000 & \textbf{63.0} \\
\textsc{sigma-dino}& 4000 & \textbf{63.0} \\
\textsc{sigma-dino}& 6000 & 62.9 \\
\bottomrule
\end{tabular}
\end{subtable}

\vspace{-0.4cm} %
\end{table*}

\begin{table*}[t]
\centering
\caption{\textbf{Ablating the projection network architecture (a) and different loss functions (b)}
Our performance is robust to our \(\varphi\) architecture. 
For deep targets, a simple L2 loss leads to a trivial solution resulting in low performance, while our \ours objective achieves the best results.} %
\label{tab:ablation_loss_combined}
\begin{subtable}{.48\linewidth}
\centering
\vspace{-1em}
\caption{Choices for $\varphi$ (see text for details of architectures)}
\begin{tabular}{cccccc}
\toprule
& Method & Architecture & Params & Acc\\
\midrule
(a)& VideoMAE & -- & -- & 56.94  \\
\midrule
\rowcolor{lightcyan}
(b)& \textsc{sigma-mlp} & base & 3M & 59.7  \\
(d)& \textsc{sigma-mlp}& shallower & 1M& 59.2  \\
(e)& \textsc{sigma-mlp}& wider &6M& 59.6 \\
\rowcolor{lightcyan}
(f)& \textsc{ sigma-dino}& DINO &22M& 63.0  \\
\bottomrule
\end{tabular}
\end{subtable}
\begin{subtable}{.48\linewidth}
\centering
\setlength{\tabcolsep}{0.3em}
\vspace{-1em}
\caption{Different reconstruction losses}
\begin{tabular}{ccccc}
\toprule
& Method & Loss & Acc \\
\midrule
(a) & VideoMAE & L2 & 56.9 \\
\midrule
(b) & \textsc{sigma-mlp} & L2 & 36.4 \\
\rowcolor{lightcyan}
(d) & \textsc{sigma-mlp}& Eq.\ref{eq:symmetric_loss} & \textbf{60.3} \\
\bottomrule
\end{tabular}
\end{subtable}%
\vspace{-0.3cm} %
\end{table*}

\section{Conclusion}
In this work, we have proposed a new self-supervised pretraining method \ours\!. 
To tackle the low semanticity of RGB-based reconstruction targets in current video modeling frameworks, we have introduced a Sinkhorn-clustering-based approach that leads to learnable and more abstract reconstruction targets. These both alleviate the issue of training collapse when simultaneously learning a projection network and can even be combined with pretrained image models, such as DINO. The resulting models outperform the state-of-the-art across a large set of datasets and benchmarks. \textit{Limitations.} Owing to our academic compute budgets, we are not able to run larger architectures, such as ViT-L or ViT-H, which require 64 GPUs~\cite{videomae}. We have, however, demonstrated positive scaling behavior for going from ViT-S to \mbox{ViT-B}.

\paragraph{Acknowledgement}
This work is financially supported by Qualcomm Technologies Inc., the University of Amsterdam, and the allowance Top consortia for Knowledge and Innovation
(TKIs) from the Netherlands Ministry of Economic Affairs
and Climate Policy.

\bibliographystyle{splncs04}
\bibliography{main}

\clearpage

\appendix

{\noindent \Huge \textbf{Supplementary Material}}

\appendix

\section{Visualization of Prototypes}

\begin{figure}[h!]
    \centering
    \includegraphics[width=\textwidth]{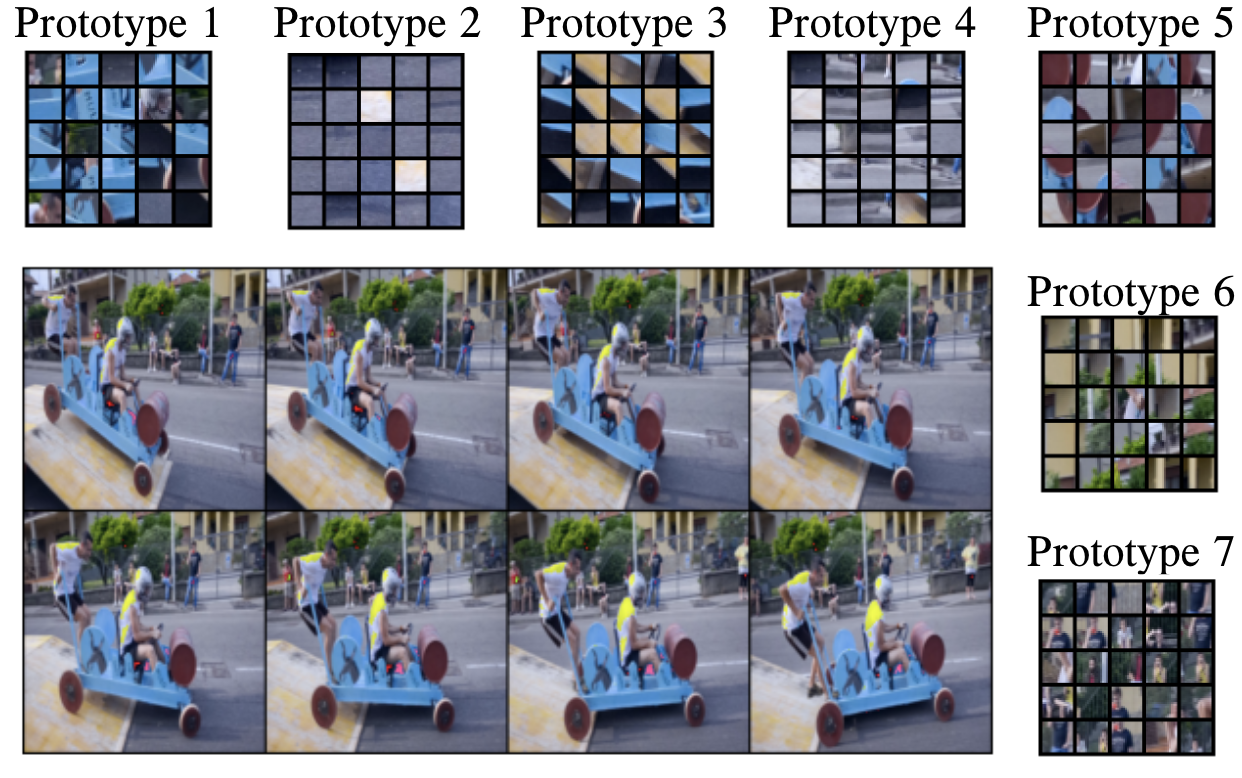}
    \caption{\textbf{Visualization of prototypes}. We visualize the 25 space-time tubes with the highest similarity to a particular prototype inside a video. For simplicity, we visualize the first patch inside the space-time tube. We observe that different prototypes attend to particular semantic parts of the video, as prototype 1 corresponds to the blue parts of the car.}
    \label{fig:prototypes}
\end{figure}

\begin{figure}[t]
    \centering
    \includegraphics[width=\textwidth]{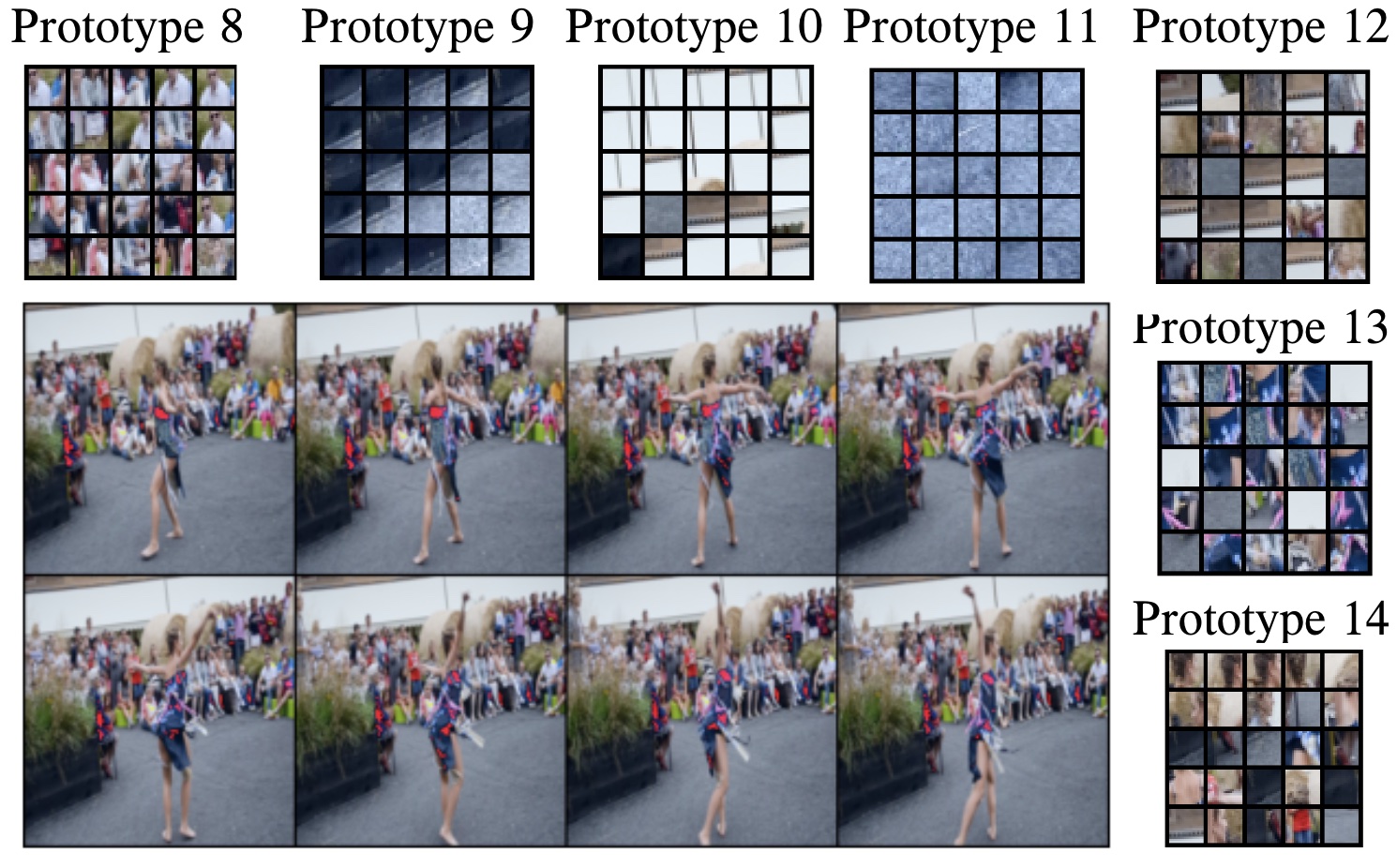}
    \caption{\textbf{Visualization of prototypes (2)}. We visualize the 25 space-time tubes with the highest similarity to a particular prototype inside a video. For simplicity, we visualize the first patch inside the space-time tube. We observe that different prototypes attend to particular semantic parts of the video, for example, prototype 1 corresponds to the person(s) in white.}
    \label{fig:prototypes2}
\vspace{-0.2cm}
\end{figure}

In this section, we analyze the prototypes learned by our method. For that, we use \ours pretrained on Kinetics and using DINO as a projection network. We visualize the 25 space-time tubes inside videos from DAVIS that have the highest similarity with a given prototype in Fig. \ref{fig:prototypes} and Fig. \ref{fig:prototypes2}. For simplcity, we visualize the first patch in time of the space-time tube. We observe in Fig. \ref{fig:prototypes} that the patches for a particular prototype are semantically similar, as prototype 6 captures the tree parts in the background, while prototype 7 captures faces/persons in the video. Similarly in Fig. \ref{fig:prototypes2} where a different set of prototypes are visualized which correspond to the white background, one to the person in white, and one to the clothing of the runner.

\section{Dataset details}
In this section, we list the details of the datasets used in our experiments.

\paragraph{{Something-Something V2 (SSv2)}~\cite{goyal2017something}} contains 220K videos with 174 action classes and is considered motion-heavy because of its focus on motion and directional aspects inherent to the actions. Example classes are: Pushing something from left to right, Pulling something from right to left, Putting something down etc.

\paragraph{Kinetics-400 (K400)\cite{kay2017kinetics}} is a dataset for recognizing actions in videos, which comprises realistic action videos gathered from YouTube. The dataset contains 306,245 short-trimmed videos, covering 400 action categories, making it one of the largest and most extensively used datasets for evaluating state-of-the-art video action recognition models. Some example classes are: Bungee jumping, Cutting pineapple, Doing aerobics.

\paragraph{DAVIS~\cite{pont20172017}} includes 150 videos, with 60 allocated for training, 30 for validation, and 60 for testing. All the validation videos for this dataset include full-frame annotations as opposed to the test set. Therefore, we use the validation split to test our object segmentation performances. 

\paragraph{YTVOS~\cite{xu2018youtube}} is a larger video object segmentation dataset compared to DAVIS, comprises 4,453 videos, each annotated under one of 65 object categories. Like DAVIS, YTVOS provides ground truth masks only for the first frames in both the test and validation sets. Therefore, a random subset of 20\% of the training set is used for evaluations. Also, the meta-information provided by datasets is used to ensure that objects within the same category have consistent class IDs.

\paragraph{UCF101~\cite{soomro2012ucf101}} The dataset comprises of 13,320 video clips that are divided into 101 categories. These 101 categories are further grouped into 5 types - Body motion, Human-human interactions, Human-object interactions, Playing musical instruments, and Sports. The combined duration of these video clips is over 27 hours. All the videos were sourced from YouTube and have a fixed frame rate of 25 FPS, with a resolution of 320 × 240. Some example classes are: Handstand Pushups, Billiards, Band Marching.

\paragraph{HMDB-51~\cite{kuehne2011hmdb}} is a dataset designed for action recognition, which has been gathered from multiple sources including movies and public databases such as the Prelinger archive, YouTube, and Google videos. The dataset contains 6,766 clips that have been categorized into 51 different action categories, each of which contains at least 100 clips. Some example classes are: Ride Horse, Shoot Gun, Turn.

\paragraph{ImageNet1K~\cite{imagenet}} is often used to train deep learning models for computer vision tasks. The ImageNet1K dataset consists of 1000 object classes, and it includes 1,281,167 training images, 50,000 validation images.

\paragraph{CIFAR-100~\cite{cifar}} is composed of 32x32 color images and includes 100 classes that are divided into 20 superclasses. Each class has 600 images. 

\paragraph{SEVERE Benchmark\cite{thoker2022severe}} encompasses eight different experimental settings from 4 different datasets. The setup for each subset in SEVERE-Benchmkark is listed in Table~\ref{tab:severe_desc}.

\begin{table*}[t]
    \centering
    \caption{\textbf{Benchmark Details} for the downstream evaluation setup, experiments, and datasets we use. For that, we use the SEVERE benchmark~\cite{thoker2022severe}.}
    \label{tab:severe_desc}
    \resizebox{\linewidth}{!}{
    \begin{tabular}{llllrrrl}
    \toprule
    \textbf{Evaluation Setup} & \textbf{Experiment} & \textbf{Dataset} & \textbf{Task} & \textbf{\#Classes} & \textbf{\#Finetuning} & \textbf{\#Testing} & \textbf{Eval Metric}\\
    \midrule

    \multirow{2}{*}{\textbf{Domain Shift}} & SSv2 & Something-Something~\cite{goyal2017something} & Action Recognition & 174 & 168,913 &  24,777 & Top-1 Accuracy\\
     & Gym99 & FineGym~\cite{shao2020finegym} & Action Recognition & 99 & 20,484 & 8,521 & Top-1 Accuracy\\
    \midrule
    \multirow{2}{*}{\textbf{Sample Efficiency}}& UCF ($10^3$) & UCF 101~\cite{soomro2012ucf101} & Action Recognition & 101 & 1,000 & 3,783 & Top-1 Accuracy\\
     & Gym ($10^3$) & FineGym~\cite{shao2020finegym} & Action Recognition & 99 & 1,000 & 8,521 & Top-1 Accuracy\\
    \midrule
    \multirow{2}{*}{\textbf{Action Granularity}} & FX-S1 & FineGym~\cite{shao2020finegym} & Action Recognition & 11 & 1,882 & 777 & Mean Class Acc\\
     & UB-S1 & FineGym~\cite{shao2020finegym} & Action Recognition & 15 & 3,511 & 1,471 & Mean Class Acc\\
    \midrule
    \multirow{2}{*}{\textbf{Task Shift}} & UCF-RC & UCFRep~\cite{ucfrep-zhang2020context} & Repetition Counting & - & 421 & 105 & Mean Error\\
     & Charades & Charades~\cite{sigurdsson2016hollywood} & Multi-label Recognition & 157 & 7,985 & 1,863 & mAP\\
    \bottomrule
    \end{tabular}}
\end{table*}

\section{Evaluation details}

\paragraph{Linear}
We follow the linear evaluation setup from MME\cite{mme_sun} and the used setup in Table~\ref{tab:linear-setting}. 

\begin{table}[h]
    \centering
    \small
    \caption{\textbf{Linear-Evaluation setting.}}
    \label{tab:app-finetune-setting}
    \begin{tabular}{l|m{1cm}<{\centering}m{1cm}<{\centering}m{1cm}<{\centering}m{1cm}<{\centering}}
         config & SSv2 & K400 & IN-1K&Others \\
         \toprule
         optimizer & \multicolumn{4}{c}{AdamW\cite{adamw}} \\
         base learning rate & \multicolumn{4}{c}{1.e-3} \\
         weight decay & \multicolumn{4}{c}{0.05} \\
         optimizer momentum & \multicolumn{4}{c}{$\beta_1,\beta_2=0.9,0.999$} \\
         layer-wise lr decay~\cite{layer_wise} & \multicolumn{4}{c}{0.75} \\
         batch size & \multicolumn{4}{c}{128}\\
         learning rate schedule & \multicolumn{4}{c}{cosine decay} \\
         training epochs & 30 & 40 & 30& 100 \\
         flip augmentation & \emph{no} & \emph{yes} & \emph{yes} & \emph{yes} \\
    \end{tabular}
    \label{tab:linear-setting}
\end{table}

\paragraph{Full finetuning}
We follow the default setup from \cite{videomae} for full finetuning and the specifics are listed in Table~\ref{tab:finetune-setting}. 

\begin{table}[h]
    \centering
    \small
    \caption{\textbf{Full finetuning evaluation setup.}}
    \label{tab:app-finetune-setting}
    \begin{tabular}{l|m{1cm}<{\centering}m{1cm}<{\centering}m{1cm}<{\centering}}
         config & SSv2 & K400 & SEVERE \\
         \toprule
         optimizer & \multicolumn{3}{c}{AdamW} \\
         base learning rate & \multicolumn{3}{c}{1.0e-3} \\
         weight decay & \multicolumn{3}{c}{0.05} \\
         optimizer momentum & \multicolumn{3}{c}{$\beta_1,\beta_2=0.9,0.999$} \\
         layer-wise lr decay\cite{layer_wise} & \multicolumn{3}{c}{0.75} \\
         batch size & 32 & 16 & 16 \\
         learning rate schedule & \multicolumn{3}{c}{cosine decay} \\
         warmup epochs & \multicolumn{3}{c}{5} \\
         training epochs & 50 & 75 & 100 \\
         flip augmentation & \emph{no} & \emph{yes} & \emph{yes} \\
         RandAug~\cite{cubuk2019randaugment} & \multicolumn{3}{c}{(9,0.5)} \\
         label smoothing\cite{szegedy2015rethinking} & \multicolumn{3}{c}{0.1} \\
         mixup~\cite{zhang2018mixup} & \multicolumn{3}{c}{0.8} \\
         cutmix~\cite{yun2019cutmix} & \multicolumn{3}{c}{1.0} \\
         drop path & \multicolumn{3}{c}{0.1} \\
    \end{tabular}
    \label{tab:finetune-setting}
\end{table}

\begin{table*}[t]
   \centering
   \caption{\textbf{Training Details} of finetuning on various downstream datasets and tasks.}
   \label{tab:finetuning}
   \resizebox{0.87\linewidth}{!}{
   \begin{tabular}{lllrrrr}
   \toprule
   \textbf{Evaluation Factor} & \textbf{Experiment} & \textbf{Dataset} & \textbf{Batch Size} & \textbf{Learning rate} & \textbf{Epochs} & \textbf{Steps} \\ %
   \midrule
   \multirow{2}{*}{\textbf{Standard}} & UCF101 & UCF 101~\cite{soomro2012ucf101} & 32&  0.0001  & 160 & [60,100,140] \\
& HMDB51 & HMDB 51~\cite{kuehne2011hmdb} & 32&  0.0001  & 160 & [60,100,140]  \\
   \midrule
   \multirow{2}{*}{\textbf{Domain Shift}} & SS-v2 & Something-Something~\cite{goyal2017something} & 32&  0.0001 & 45  & [25, 35, 40] \\
    & Gym-99 & FineGym~\cite{shao2020finegym} & 32&  0.0001 & 160 & [60,100,140] \\
   \midrule
   \multirow{2}{*}{\textbf{Sample Efficiency}}& UCF ($10^3$) & UCF 101~\cite{soomro2012ucf101} &  32&  0.0001 & 160 & [80,120,140]  \\
    & Gym ($10^3$) & FineGym~\cite{shao2020finegym} & 32&  0.0001 & 160 & [80,120,140]  \\
   \midrule
   \multirow{2}{*}{\textbf{Action Granularity}} & FX-S1 & FineGym~\cite{shao2020finegym} & 32&  0.0001 & 160 & [70,120,140]  \\
    & UB-S1 & FineGym~\cite{shao2020finegym} & 32&  0.0001 & 160 & [70,120,140]  \\
   \midrule
   \multirow{2}{*}{\textbf{Task Shift}} & UCF-RC & UCFRep~\cite{ucfrep-zhang2020context} & 32&  0.00005 & 100 & -  \\
    & Charades & Charades~\cite{sigurdsson2016hollywood} &  16&  0.0375 & 57 & [41,49]  \\
   \bottomrule
   \end{tabular}}
\end{table*}

\paragraph{Unsupervised segmentation} We obtain video clips of size $[T, 3, 224, 224]$ from two datasets: DAVIS~\cite{pont20172017} and YTVOS~\cite{xu2018youtube}. For DAVIS~\cite{pont20172017}, we use clips of length $T=16$, and for YTVOS~\cite{xu2018youtube}, the length is $T=4$. Each clip is paired with its corresponding ground truth and fed into the model to extract final dense features of size $[\frac{T}{2}, d, 14, 14]$, where $d$ represents the dimension of the encoder. 

Next, we resize the ground truth and feature maps to size 28 using nearest neighbor and linear interpolation, respectively. We then cluster the feature maps with different granularity values of $K$. We set $K$ to the ground truth object counts for clustering, and three times higher than the average object counts per clip, which is 6, for overclustering evaluations. This results in clustering and overclustering maps. Finally, we repeat every cluster map two times and group the clusters into ground-truth classes by matching them either by pixel-wise precision or Hungarian matching on merged cluster maps, similar to~\cite{salehi2023time}.

\paragraph{Severe benchmark}

For action recognition tasks in SEVERE benchmark (\textbf{GYM99}, \textbf{UCF}, \textbf{FX-S1} and \textbf{UBS1}) we follow the finetuning setup from Table~\ref{tab:finetune-setting}. 

For the Repetition Counting Task (denoted as \textbf{UCF-RC}), we adhere to the implementation details specified in the original work~\cite{ucfrep-zhang2020context} on repetition counting. From the annotated video dataset, we construct 2 million sequences, each consisting of 32 frames with a spatial resolution of $224 \times 224$ pixels. These sequences serve as the input to our model. The training process spans over 100 epochs with a batch size of 32, utilizing the Adam\cite{kingma2017adam} optimizer. The learning rate is set to $5 \times 10^{-5}$. For the evaluation phase, we follow ~\cite{ucfrep-zhang2020context} to report the mean counting error.

For the Multi-label classification on \textbf{Charades}, we employ \cite{large-scale-feichtenhofer2021large} to incorporate a per-class sigmoid output layer for multi-class prediction. In the training phase, we sample 16 frames with a stride of 8 from each video. The frames are resized to a spatial resolution of $224 \times 224$ pixels. We apply several data augmentation techniques, including random short-side scaling, random spatial cropping, and horizontal flipping. The model is trained over 57 epochs, utilizing a batch size of 16 and a learning rate of $1 \times 10^{-4}$. For the evaluation phase, spatiotemporal max-pooling is executed over 10 distinct clips from each video to aggregate the predictions. The performance is quantified using the mean Average Precision (mAP) across all classes.

\section{Extended comparison for full finetuning on SSv2 and K400}
We provide an extended version of comparison with state-of-the-art for full finetuning in Tab.~\ref{tab:ext_k400} and Tab.~\ref{tab:ext_ssv2}. As is shown, \ours improves VideoMAE~\cite{tong2022videomae} baseline by 0.8\% and 1.3\% on SSv2(Tab.~\ref{tab:ext_ssv2}) when pretrained on K400 or SSv2 using an MLP projection network.
Using a pretrained DINO~\cite{caron2021emerging} model as the projection network results in even larger improvement, reaching 1\% and 1.8\% across different pretrainings, getting state-of-the-art results across the models trained for the same number of epochs. For K400 we observe similar results. As shown in Tab.~\ref{tab:ext_k400}, \ours considerably improves VideoMAE~\cite{tong2022videomae} baseline and sets a new state-of-the-art for 800 training epochs. 

MVD~\cite{wang2023masked} achieves good performance while using a computationally expensive approach requiring longer pretraining. First, a video model, following VideoMAE~\cite{tong2022videomae}, is trained on K400 for 1600 epochs. Then, an image model, following MAE~\cite{he2022masked}, is trained on ImageNet~\cite{imagenet} for 1600 epochs. Finally, the VideoMAE and MAE models are kept frozen and serve as the teachers for the main video model which is trained via distillation for 400 epochs. This complex and multi-step training process makes it hard to provide a one-to-one comparison between this and other methods. MGM~\cite{guided_masking_fan} and MME~\cite{mme_sun} are two other models that have been trained with a higher number of epochs, yet they still perform comparably to our model which was trained with half the number of epochs, based on the K400 benchmark.

\begin{table*}[t]
\centering
\caption{\textbf{Benchmark I: Comparison for full finetuning on Kinetics 400} (K400). We compare against all previous methods for pretraining the ViT-Base backbone on K400 and subsequently, fully finetuning the backbone with the K400 labels. M. Guid. denotes motion guidance such as optical flow used e.g. reconstructing targets or masking.}
\resizebox{1\textwidth}{!}{%
\begin{tabular}{ll cccccccc}
\toprule
&Method & M. Guid. & Backbone & Epochs & Extra data & Frames & Params & Top-1 \\
\midrule
&\textbf{\textit{supervised}} & & & & & & & & \\
&\quad SlowFast~\cite{Feichtenhofer2018SlowFastNF}  & --& ResNet101 & -- & -- & -- & 16+64 & 60 & 79.8  \\
&\quad MViTv1~\cite{mvit_v1_fan} & --& MViTv1-B & -- & -- & 32  & 37 & 80.2 \\
&\quad TimeSformer~\cite{Bertasius2021IsSA} & --& ViT-B & -- & IN-21K & 96 & 430 & 80.7\\
&\quad VideoSwin~\cite{videoswin_liu} & -- & Swin-L & -- & IN-21K & 32& 197 & 83.1 \\
\midrule
&\textbf{\textit{self-supervised}} & & & & & & & & \\
&\quad \color{gray}VideoMAE~\cite{videomae}& $\times $&\color{gray}ViT-S  & \color{gray}1600 &\color{gray}-- & \color{gray}16 & \color{gray}87 & \color{gray}79.0 \\
&\quad \color{gray}MVD~\cite{mvd_wang} & $\times $&\color{gray}ViT-S  & \color{gray}1600 + 400 & \color{gray}IN-1K &\color{gray}16 & \color{gray}22 & \color{gray}\text{80.6} \\
&\quad \cellcolor{lightcyan}\textsc{sigma-dino} (ours)  & \cellcolor{lightcyan}$\times $&\cellcolor{lightcyan}ViT-S  & \cellcolor{lightcyan}800& \cellcolor{lightcyan}IN-1K &\cellcolor{lightcyan}16 &\cellcolor{lightcyan} 87 & \cellcolor{lightcyan} {79.4} \\
\cmidrule{3-9}
&\quad \color{gray}VIMPAC~\cite{vimpac}& $\times $&\color{gray}ViT-L & \color{gray}100 & \color{gray}HowTo100M+DALL-E & \color{gray}10  & \color{gray}307 & \color{gray}77.4 \\
&\quad VideoMAE~\cite{videomae} &$\times $& ViT-B  & 800 & -- & 16  & 87 &  80.0 \\
\parbox[t]{3mm}{\multirow{5}{*}{\rotatebox[origin=c]{90}{\textbf{K400 pretraining}}}}&\quad \color{gray}VideoMAE~\cite{videomae}& $\times$& \color{gray}ViT-B  & \color{gray}1600 & \color{gray}-- &\color{gray}16  & \color{gray}87 & \color{gray}80.9 \\
&\quad OmniMAE~\cite{girdhar2023omnimae} & $\times $&ViT-B & 800 & IN-1K & 16& 87 & 80.8 \\
&\quad \color{gray}ST-MAE~\cite{feichtenhofer2022masked}& $\times $&\color{gray}ViT-B  & \color{gray}1600 & \color{gray}-- & \color{gray}16  & \color{gray}87 & \color{gray}81.3 \\
&\quad \color{gray}MME~\cite{mme_sun}& $\checkmark $&\color{gray}ViT-B  & \color{gray}1600 & \color{gray}-- & \color{gray}16  &\color{gray}87 & \color{gray}81.8 \\
&\quad \color{gray}MVD~\cite{mvd_wang} & $\times $&\color{gray}ViT-B  & \color{gray}1600 + 400 & \color{gray}IN-1K &\color{gray}16& \color{gray}87 & \color{gray}\text{82.7} \\
&\quad CMAE-V~\cite{cmae_v_lu} & $\times$& ViT-B  & 800 &  -- & 16  & 87 & 80.2 \\
&\quad \color{gray}CMAE-V~\cite{cmae_v_lu} & $\times $& \color{gray}ViT-B  & \color{gray}1600 & \color{gray}-- & \color{gray}16 & \color{gray}87 & \color{gray}80.9 \\
&\quad MGM~\cite{guided_masking_fan}& $\checkmark$ & ViT-B  & 800 & -- & 16& 87 &  {80.8} \\
&\quad \color{gray}MGM~\cite{guided_masking_fan}& \color{gray}$\checkmark $&\color{gray}ViT-B & \color{gray}1600  & \color{gray}-- &\color{gray}16& \color{gray}87 & \color{gray}81.7 \\
&\quad MGMAE~\cite{mgmae_huang}& $\checkmark $& ViT-B & 800  & -- &16& 87 &81.2 \\
&\quad \cellcolor{lightcyan}\textsc{sigma-mlp}  (ours)& \cellcolor{lightcyan}$\times $&\cellcolor{lightcyan}ViT-B  & \cellcolor{lightcyan}800& \cellcolor{lightcyan}-- &\cellcolor{lightcyan}16& \cellcolor{lightcyan}87 & \cellcolor{lightcyan}80.2 \\
&\quad \cellcolor{lightcyan}\textsc{sigma-dino} (ours)&\cellcolor{lightcyan}$\times$ & \cellcolor{lightcyan}ViT-B  &\cellcolor{lightcyan}800&\cellcolor{lightcyan} IN-1K &\cellcolor{lightcyan}16 & \cellcolor{lightcyan}87 & \cellcolor{lightcyan}{\textbf{81.6}}\\
\bottomrule
\end{tabular}%
}

\label{tab:ext_k400}
\end{table*}

\begin{table*}[t]
\centering
\caption{\textbf{Benchmark I: Comparison for full finetuning on Something-Something V2} (SSv2). The top part compromises supervised methods while the remaining methods are pretrained in a self-supervised manner. The middle section evaluates models trained on Kinetics 400 (K400) data for pretraining whereas the bottom part mainly uses SSv2 data. We compare against all previous methods pretrained on the ViT-Base backbone. M. Guid. denotes motion guidance such as optical flow used e.g. reconstructing targets or masking.}
\resizebox{1\textwidth}{!}{%
\begin{tabular}{ll cccccccc}
\toprule
&Method & M. Guid. & Backbone & Epochs & Extra data & Frames  & Params & Top-1 \\
\midrule
&\textbf{\textit{supervised baselines}} & & & & & & & & \\
&\quad SlowFast\cite{Feichtenhofer2018SlowFastNF}  & $\times$ & ResNet101 & -- & K400 & 8+32  & 53 & 63.1 \\
&\quad MViTv1~\cite{mvit_v1_fan} & $\times$ & MViTv1-B & -- & K400 & 64 & 37 & 67.7 \\
&\quad TimeSformer~\cite{Bertasius2021IsSA} & $\times$ & ViT-B & -- & IN-21K & 8 & 121 & 59.5 \\
&\quad VideoSwin~\cite{videoswin_liu} & $\times$ & Swin-B & -- & IN-21K & 32 & 88 & 69.6 \\
\midrule
&\textbf{\textit{self-supervised}} & & & & & & & & \\
&\quad \color{gray}MVD~\cite{mvd_wang}  & $\times$ &  \color{gray}ViT-S  & \color{gray}1600 + 400 & \color{gray}IN-1K+K400 &\color{gray}16 & \color{gray}22 & \color{gray}70.7 \\
&\quad \cellcolor{lightcyan}\textsc{sigma-dino} (ours) & \cellcolor{lightcyan} $\times$ & \cellcolor{lightcyan}ViT-S  & \cellcolor{lightcyan}800& \cellcolor{lightcyan}IN-1K+K400 &\cellcolor{lightcyan}16& \cellcolor{lightcyan}25 & \cellcolor{lightcyan} 68.7\\
\cmidrule{3-9}
\parbox[t]{3mm}{\multirow{5}{*}{\rotatebox[origin=c]{90}{\textbf{K400 pretraining}}}}&\quad \color{gray}BEVT~\cite{wang2022bevt} & $\times$ & \color{gray}Swin-B & \color{gray}150 & \color{gray}IN-1K+K400 & \color{gray}32 & \color{gray}88 & \color{gray}67.6 \\
&\quad \color{gray}BEVT~\cite{wang2022bevt} & $\times$ & \color{gray}Swin-B & \color{gray}150 & \color{gray}IN-1K+K400+DALL-E & \color{gray}32& \color{gray}88 & \color{gray}70.6 \\
&\quad \color{gray}VIMPAC~\cite{vimpac} & $\times$ & \color{gray}ViT-L & \color{gray}100 & \color{gray}HowTo100M+DALL-E & \color{gray}10 & \color{gray}307 & \color{gray}68.1 \\
&\quad OmniMAE~\cite{girdhar2023omnimae} & $\times$ & ViT-B & 800 & IN-1K+K400 &16&  86 & 69.0 \\
&\quad VideoMAE~\cite{videomae} & $\times$ & ViT-B  & 800 & K400 & 16 & 87 & 68.5 \\
&\quad \color{gray}VideoMAE~\cite{videomae} & $\times$ & \color{gray}ViT-B  & \color{gray}2400 & \color{gray}K400 & \color{gray}16  & \color{gray}87 & \color{gray}69.7 \\
&\quad MME~\cite{mme_sun} & $\checkmark$ & ViT-B  & 800 & K400 & 16 & 87 & {70.5} \\
&\quad \color{gray}MVD~\cite{mvd_wang} & $\times$ & \color{gray}ViT-B  & \color{gray}1600 + 400 & \color{gray}IN-1K+K400 &\color{gray}16& \color{gray}87 & \color{gray}72.5 \\
&\quad \cellcolor{lightcyan}\textsc{sigma-mlp} (ours) & \cellcolor{lightcyan} $\times$&  \cellcolor{lightcyan}ViT-B  & \cellcolor{lightcyan}800& \cellcolor{lightcyan}K400&\cellcolor{lightcyan}16& \cellcolor{lightcyan}90 & \cellcolor{lightcyan}69.8 \\
&\quad \cellcolor{lightcyan}\textsc{sigma-dino} (ours) & \cellcolor{lightcyan} $\times$ & \cellcolor{lightcyan}ViT-B  & \cellcolor{lightcyan}800& \cellcolor{lightcyan}IN-1K+K400 &\cellcolor{lightcyan}16& \cellcolor{lightcyan}87 & \cellcolor{lightcyan}\textbf{71.1} \\
\cmidrule{2-9}
&\quad VideoMAE~\cite{videomae} & $\times$& ViT-S  & 2400 & -- & 16  & 22 & 66.8 \\
&\quad \cellcolor{lightcyan}\textsc{sigma-dino} (ours)&  \cellcolor{lightcyan} $\times$ & \cellcolor{lightcyan}ViT-S  & \cellcolor{lightcyan}2400& \cellcolor{lightcyan}IN-1K &\cellcolor{lightcyan}16 & \cellcolor{lightcyan}22 & \cellcolor{lightcyan}\textbf{68.6} \\
\cmidrule{3-9}
&\quad OmniMAE~\cite{girdhar2023omnimae} & $\times$ & ViT-B & 800 & IN-1K & 16& 86 & 69.5 \\
&\quad VideoMAE~\cite{videomae}& $\times$& ViT-B & 800 &-- & 16  & 87 & 69.6\\
\parbox[t]{3mm}{\multirow{5}{*}{\rotatebox[origin=c]{90}{\textbf{SSv2 pretraining}}}} &\quad \color{gray}VideoMAE~\cite{videomae}& $\times$& \color{gray}ViT-B & \color{gray}2400 &\color{gray}-- & \color{gray}16 & \color{gray}87 & \color{gray}{70.8}\\
&\quad CMAE-V~\cite{cmae_v_lu}& $\times$ & ViT-B  & 800 &  -- & 16 & 87 & 69.7 \\
&\quad \color{gray}CMAE-V~\cite{cmae_v_lu}& $\times$ & \color{gray}ViT-B  & \color{gray}1600 & \color{gray}-- & \color{gray}16 & \color{gray}87 & \color{gray}70.5 \\

&\quad \cellcolor{lightcyan}\textsc{sigma-mlp} (ours) & \cellcolor{lightcyan} $\times$ & \cellcolor{lightcyan}ViT-B  & \cellcolor{lightcyan}800& \cellcolor{lightcyan}-- &\cellcolor{lightcyan}16 & \cellcolor{lightcyan}87 &\cellcolor{lightcyan}{70.4} \\
&\quad \cellcolor{lightcyan}\textsc{sigma-dino} (ours) & \cellcolor{lightcyan} $\times$ & \cellcolor{lightcyan}ViT-B  & \cellcolor{lightcyan}800& \cellcolor{lightcyan}IN-1K &\cellcolor{lightcyan}16 & \cellcolor{lightcyan}87 & \cellcolor{lightcyan}\textbf{70.9} \\
\cmidrule{3-9}
&\quad MME~\cite{mme_sun}& $\checkmark$& ViT-B  & 800 & -- & 16& 87 & 70.0 \\
&\quad MGM~\cite{guided_masking_fan}& $\checkmark$& ViT-B   & 800 & --  &16& 87 &  70.6 \\
&\quad \color{gray}MGM~\cite{guided_masking_fan}& $\checkmark$ & \color{gray}ViT-B & \color{gray}1200 & \color{gray}-- & \color{gray}16& \color{gray}87 & \color{gray}\text{71.6} \\
&\quad \color{gray}MGM~\cite{guided_masking_fan}& $\checkmark$& \color{gray}ViT-B & \color{gray}1600  & \color{gray}--&\color{gray}16& \color{gray}87 & \color{gray}71.8 \\
&\quad MGMAE~\cite{mgmae_huang} & $\checkmark$ & ViT-B & 800  & \color{gray}--&16& 87 & 71.0 \\
&\quad \cellcolor{lightcyan}\textsc{sigma-mlp} (ours) & \cellcolor{lightcyan} $\checkmark$ & \cellcolor{lightcyan}ViT-B  & \cellcolor{lightcyan}800& \cellcolor{lightcyan}-- &\cellcolor{lightcyan}16 & \cellcolor{lightcyan}87 &\cellcolor{lightcyan}\textbf{71.2} \\
&\quad \cellcolor{lightcyan}\textsc{sigma-dino} (ours) & \cellcolor{lightcyan} $\checkmark$ & \cellcolor{lightcyan}ViT-B  & \cellcolor{lightcyan}800& \cellcolor{lightcyan}IN-1K &\cellcolor{lightcyan}16 & \cellcolor{lightcyan}87 & \cellcolor{lightcyan}\textbf{71.2} \\

\bottomrule
\end{tabular}%
}

\label{tab:ext_ssv2}
\end{table*}

\end{document}